\documentclass[lettersize,journal]{IEEEtran}
\usepackage{amsmath,amsfonts}
\usepackage{array}
\usepackage{textcomp}
\usepackage{stfloats}
\usepackage{url}
\usepackage{verbatim}
\usepackage{graphicx}
\def\BibTeX{{\rm B\kern-.05em{\sc i\kern-.025em b}\kern-.08em
		T\kern-.1667em\lower.7ex\hbox{E}\kern-.125emX}}
\usepackage{balance}
\usepackage{subfigure}
\usepackage[colorlinks=true, linkcolor=blue, citecolor=blue, urlcolor=blue]{hyperref}
\usepackage{tabularx}  

\usepackage{algorithm}
\usepackage{algpseudocode}
\begin{document}
	\title{LSTP-Nav: Lightweight Spatiotemporal Policy for Map-free Multi-agent Navigation with LiDAR\\
		\thanks{Project page: \url{https://sites.google.com/view/xingrong2024efficient}.}
		\thanks{This work is supported by National Natural Science Foundation of China under Grant 62473191, Shenzhen Key Laboratory of Robotics Perception and Intelligence (ZDSYS20200810171800001), Shenzhen Science and Technology Program under Grant 20231115141459001, Guangdong Basic and Applied Basic Research Foundation under Grant 2025A1515012998, and the High level of special funds (G03034K003) from Southern University of Science and Technology, Shenzhen, China. (Corresponding author: Jiankun Wang)}
		
		\thanks{Xingrong Diao, Zhirui Sun, Jianwei Peng, Jiankun Wang are with Shenzhen Key Laboratory of Robotics Perception and Intelligence, Department of Electronic and Electrical Engineering, Southern University of Science and Technology, Shenzhen, China (e-mail: wangjk@sustech.edu.cn).}
		
		\thanks{Zhirui Sun and Jiankun Wang are also with Jiaxing Research Institute, Southern University of Science and Technology, Jiaxing, China.}
		}
	
	\author{Xingrong Diao, Zhirui Sun, Jianwei Peng, Jiankun Wang*}
	
	\markboth{Journal of \LaTeX\ Class Files,~Vol.~18, No.~9, September~2020}%
	{How to Use the IEEEtran \LaTeX \ Templates}
	
	\maketitle
	
	\begin{abstract}
		Safe and efficient multi-agent navigation in dynamic environments remains inherently challenging, particularly when real-time decision-making is required on resource-constrained platforms. Ensuring collision-free trajectories while adapting to uncertainties without relying on pre-built maps further complicates real-world deployment. To address these challenges, we propose LSTP-Nav, a lightweight end-to-end policy for multi-agent navigation that enables map-free collision avoidance in complex environments by directly mapping raw LiDAR point clouds to motion commands. At the core of this framework lies LSTP-Net, an efficient network that processes raw LiDAR data using a GRU architecture, enhanced with attention mechanisms to dynamically focus on critical environmental features while minimizing computational overhead. Additionally, a novel HS reward optimizes collision avoidance by incorporating angular velocity, prioritizing obstacles along the predicted heading, and enhancing training stability. To narrow the sim-to-real gap, we develop PhysReplay-Simlab, a physics-realistic multi-agent simulator, employs localized replay to mine near-failure experiences. Relying solely on LiDA, LSTP-Nav achieves efficient zero-shot sim-to-real transfer on a CPU-only robotic platform, enabling robust navigation in dynamic environments while maintaining computation frequencies above 40 Hz. Extensive experiments demonstrate that LSTP-Nav outperforms baselines with a 9.58\% higher success rate and a 12.30\% lower collision rate, underscoring its practicality and robustness for real-world applications.
	\end{abstract}
	
	\begin{IEEEkeywords}
		Deep reinforcement learning, multi-agent, collision avoidance, map-free navigation, sim-to-real transfer.
	\end{IEEEkeywords}
	
	\section{Introduction}
	Multi-agent systems are transforming robotic applications, ranging from automotive assembly lines requiring millimeter-precision coordination \cite{Industrial} to airport robots autonomously routing hundreds of airport luggage carts \cite{Xie}. Core to these advances is developing navigation algorithms that guarantee collision-free trajectories in dense scenarios, deadlock prevention during high-stakes coordination, and real-time adaptation to environmental uncertainties.
	
	Multi-agent navigation approaches are fundamentally classified along two dimensions: system architecture and algorithmic paradigm. From a system architecture perspective, conventional designs are categorized as centralized or decentralized. In the centralized architecture, a central server uses global information to plan trajectories for all agents, leading to challenges in scalability (computational cost increases with agent numbers) and robustness (single point of failure) \cite{TEA*}. The decentralized architecture, implemented by methods such as Reciprocal Velocity Obstacle (RVO) \cite{RVO} and Optimal Reciprocal Collision Avoidance (ORCA) \cite{ORCA}, allows agents to plan independently based on local observations. However, these methods often depend on unrealistic assumptions of perfect perception, face difficulties in scalability and complex environments, and require careful parameter tuning \cite{FIMVO}.
	
	Complementary to this architectural classification is the algorithmic paradigm, which distinguishes rule-based and optimization-based approaches. Rule-based variants like AVOCADO \cite{martinez2025avocado} (leveraging opinion dynamics for adaptive cooperation) and V-RVO \cite{V-RVO} (integrating Buffered Voronoi Cells for clearance) mitigate specific issues like deadlocks but retain sensitivity to scenario-specific configurations, limiting practical robustness. Optimization-based approaches frame navigation as an explicit optimization problem over paths or velocities. Established techniques include Model Predictive Control (MPC) \cite{MPC, peng2023mpc} and the Dynamic Window Approach (DWA) \cite{DWA, dobrevski2024dynamic}, etc. Recent advances integrate sophisticated frameworks like safe corridors and Graphs of Convex Sets (GCS) to enhance reliability and handle complex constraints. For instance, Marcucci et al. \cite{GCS} propose a GCS-based framework enabling obstacle-aware trajectory planning via tractable convex relaxation. While capable of generating theoretically optimal trajectories, these methods often incur high computational costs or face significant challenges in solving non-convex optimization problems, particularly within large-scale or unstructured environments, limiting their applicability.
	
	To address the limitations of conventional methods, particularly in handling uncertainty, generalization, and complex interactions, learning-based navigation has been developed. Imitation learning (IL) approaches \cite{chen2024imitation, Mapless} train policies by mimicking expert demonstrations and mapping environmental observations to actions. A specific example is that VOILA \cite{VOILA} derives vision-based navigation policies from a single physical agent demonstration. Supervised methods like RASTA \cite{LSTNE} utilize risk-aware instance weighting and pseudo-labeling for self-supervised learning in high-risk terrains. However, IL and supervised approaches are fundamentally constrained by their dependence on expert knowledge and the quality and diversity of the training data, inherently limiting their generalization capability. 
	
	Reinforcement learning (RL) offers an alternative paradigm, learning optimal navigation policies through reward-driven environmental interactions. The performance of RL policies critically depends on reward design, interaction quality, and policy architecture. For instance, DRL-VO \cite{DRL-VO} introduces a velocity obstacle term into the reward function, enabling mobile robots to navigate autonomously in spaces filled with static obstacles and dense pedestrian traffic; CAN \cite{CAN} provides the agent with a curriculum and a replay buffer of clipped collision segments, enhancing performance in obstacle-rich environments; CNNs process perceptual inputs \cite{kalidas2023deep}, while recurrent structures leverage historical states \cite{RNN-based}. Due to safety constraints, RL training predominantly occurs in simulation. Narrowing the critical sim-to-real gap is addressed through techniques like progressive networks \cite{chukwurah2024sim}, domain randomization \cite{s2r-RL}, cyclic-consistent GANs \cite{S2RLF}, and hybrid control integration \cite{IJRR}, improving real-world transferability, though challenges remain \cite{BADGR}.
	
	\begin{figure*}[t]  
		\centering  
		\includegraphics[width=\textwidth]{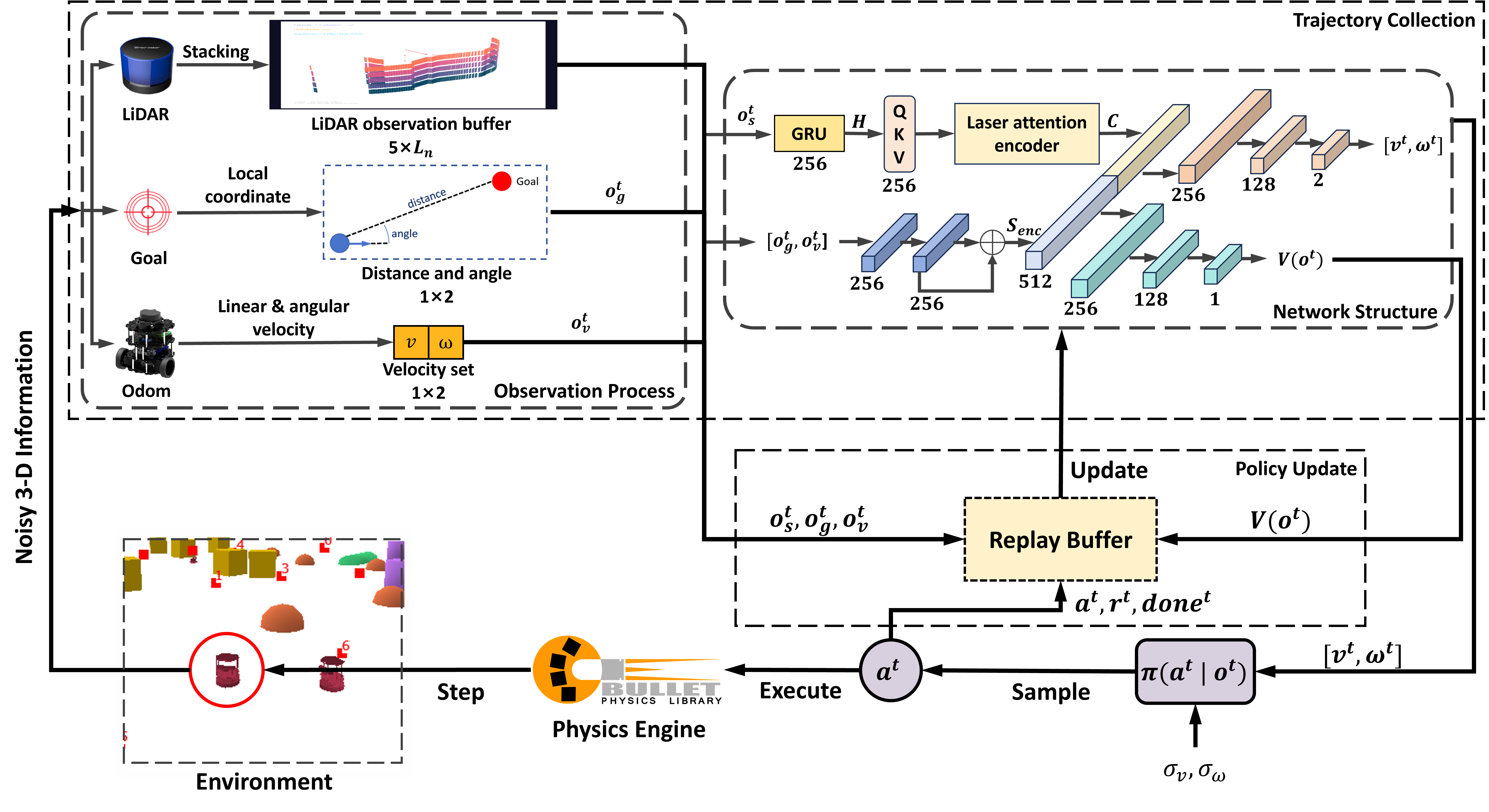}  
		\caption{The overall framework diagram depicts the proposed approaches, which comprise trajectory collection and policy optimization phases. During the trajectory collection phase, agents independently execute actions. The environment returns rewards and next-step observations, which are stored as per-agent transitions in the replay buffer. During the policy optimization phase, the trajectories from the replay buffer compute policy loss. The optimizer then updates network parameters via gradient descent.}
		\label{policy2}
	\end{figure*} 
	
	In this paper, we present \textbf{LSTP-Nav (Lightweight End-to-End Spatiotemporal Policy for Navigation)}, a decentralized end-to-end policy that efficiently maps raw 2D LiDAR inputs to motion commands for map-free multi-agent navigation. At its core, the \textbf{Lightweight Spatiotemporal Policy Network (LSTP-Net)} uses a gated recurrent unit-based architecture with attention mechanisms to process raw LiDAR data, dynamically focusing on critical features, addressing partial observability, and mitigating suboptimal behaviors like freezing at narrow passages, all while maintaining low computational overhead. LiDAR is selected as the primary sensing modality for two key reasons \cite{review}: (1) its ability to significantly narrow the sim-to-real gap compared to other sensor types, and (2) its compact data representation, enabling efficient policy transfer. Additionally, a novel \textbf{heading stability reward (HS reward)} is introduced to optimize collision avoidance during interactions. To facilitate realistic training, we also develop \textbf{PhysReplay-Simlab}, a physics-enhanced simulator designed specifically for multi-agent navigation tasks. It incorporates realistic physics, and a localized replay mechanism that actively mines near-failure experiences to improve learning efficiency. Relying exclusively on LiDAR data, LSTP-Nav directly generates control commands for safe, real-time, map-free navigation of resource-constrained robots in dynamic and unstructured environments.
	
	Our main contributions are summarized as follows:
	
	\begin{enumerate}
		
		\item We propose \textbf{LSTP-Nav}, a lightweight end-to-end spatiotemporal policy for map-free multi-agent navigation. It leverages the \textbf{LSTP-Net} to process raw LiDAR data with minimal computational overhead, and introduces a novel \textbf{HS reward} to enhance obstacle avoidance.
		
		\item We develop \textbf{PhysReplay-Simlab}, a dedicated simulator for multi-agent systems, with a replay mechanism, enhancing training robustness and narrowing the sim-to-real gap.
		
		\item We deploy LSTP-Nav in real-time on a CPU-only robotic platform, achieving zero-shot sim-to-real transfer while maintaining computation frequencies above 40 Hz. Both LSTP-Nav and PhysReplay-Simlab are also available as open-source packages on GitHub.
		
	\end{enumerate}
	
	\section{Methodology} 
	\label{Problem formulation}
	In this section, we will introduce the problem statement, the PhysReplay-Simlab, and the LSTP-Nav framework.
	
	\subsection{Problem Statement}
	This work models the multi-agent navigation process as a Partially Observable Markov Decision Process (POMDP). The POMDP can be defined as $<\mathcal{I}, \mathcal{S},\mathcal{O},\mathcal{A},\mathcal{R},\mathcal{T}, done, \gamma>$, where $\mathcal{I}=\{1,..., N\}$ denotes the set of $N$ agents, $\mathcal{S}$ denotes the set of state of the environment, $\mathcal{O}$ denotes the set of observable states of agents, $\mathcal{A}$ denotes the joint actions, $\mathcal{R}$ denotes the reward function, $\mathcal{T}:\mathcal{S}\times\mathcal{A}\mapsto\mathcal{S}$ denotes the transition function, $done$ is the termination signal, and $\gamma\in [0,1)$ denotes the discount factor. An episode refers to a full agent-environment interaction cycle. An epoch denotes a policy parameter update step. An iteration encompasses one complete training cycle, typically consisting of multiple episodes and epochs. Batches of episodes terminate when reaching the maximum timestep threshold $T_{max}$, irrespective of task completion status.
	
	Specifically, the system state evolves according to $\mathcal{T}(s^{t+1}|s^t, \mathcal{A})$, with observable state following $\mathcal{T}(\mathcal{O}^{t+1}|\mathcal{O}^t,\mathcal{A})$. The joint reward function is $\mathcal{R}=\{r(s_1^t,a_1^t), ..., r(s_N^t,a_N^t)\}$. Each agent $i$ collects transitions $<o_i^t,a_i^t,r_i^t,o_i^{t+1},done^t, V(o^t)>$ over $T_{max}$ timesteps, aiming to maximize its discounted return $G_i^T=\sum_{t=1}^T\gamma^tr(s_i^t, a_i^t)$. The state-value function $V^\pi(s_i^T)=\mathbb{E}[G_i^T|s^1_i]$ estimates expected return, while the action-value function $Q^\pi(s_i^T,a_i^T)=\mathbb{E}[G_i^T+\gamma V^\pi(s_i^T)|s^1_i,a^1_i]$ evaluates state-action pairs. Action quality is measured by the advantage function $A^\pi=Q^\pi-V^\pi$.
	
	As illustrated in Fig. \ref{policy2}, the training framework operates through iterations where agents navigate unknown environments and independently collect trajectories. Each iteration alternates between data collection and policy optimization until convergence or a maximum iterations. During the data collection phase, agents perform $T_{max}$ episodes where they: (1) receive observations $o_s^t$, $o_g^t$, $o_v^t$; (2) compute action distributions and state values $v^t$ using $\pi_\theta$ and $V_\phi$; (3) sample and execute actions $a^t$; and (4) store transitions $<o^t, a^t, r^t, o^{t+1}, done^t, V_\phi(o^t)>$ with simulator feedback in a replay buffer. The subsequent optimization phase minimizes the training loss via $K$ epochs of minibatch stochastic gradient descent. Finally, the replay buffer is reset for the next iteration.
	
	\textit{Observation and action space:} The observation vector $o^t=[o_s^t,o_g^t,o_v^t]$ includes stacked LiDAR measurements $o_s^t\in\mathbb{R}^{5\times N_{laser}}$ from 5 consecutive frames (using $N_{laser}=130$ beams with 4m range), goal information $o_g^t=[d_g^t, \psi_g^t]\in\mathbb{R}^2$ (distance $d_g^t$ clipped to $(0,4]$m for scale-invariant learning and relative heading $\psi_g^t$), and velocity information $o_v^t=[v_i^t, \omega_i^t]\in\mathbb{R}^2$ (linear velocity $v_i^t \in [0, 1]$ m/s and angular velocity $\omega_i^t \in [-\pi,\pi]$ rad/s). These inputs are processed by a policy network $\pi_\theta$ that outputs parameters of a Gaussian distribution, from which actions $a^t_i=[v^t_i,\omega^t_i]$. The control frequency of agent is set at 60 Hz for precise actuation. Simultaneously, a value network $V_\phi$ estimates state values to guide policy updates.
	
	\subsection{PhysReplay-Simlab}
\begin{figure}[htbp]  
		\centering  
		\subfigure[2D simulator]{\includegraphics[height=0.17\textwidth]{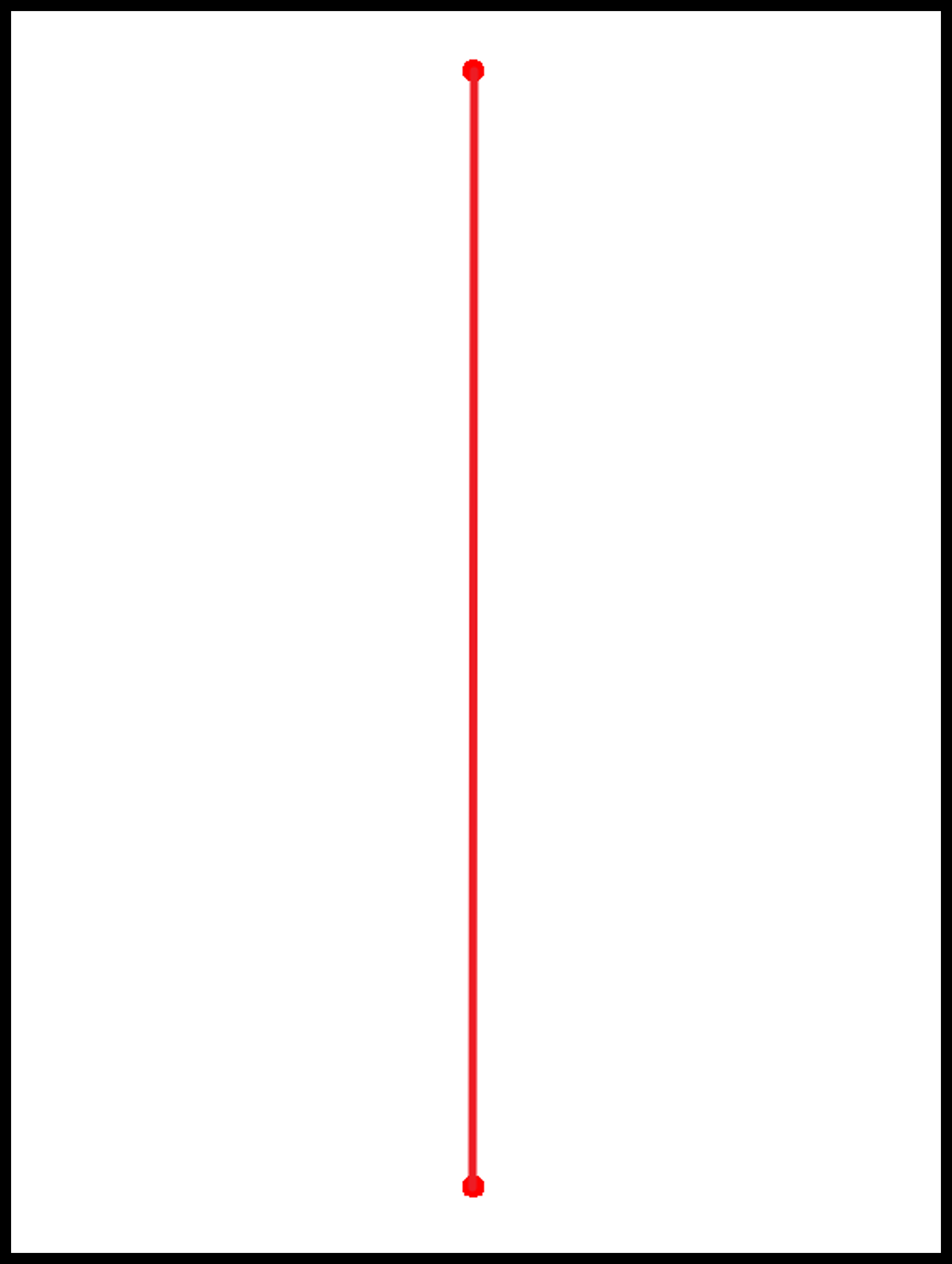}}
		\hfill 
		\subfigure[Physical based simulator]{\includegraphics[height=0.17\textwidth]{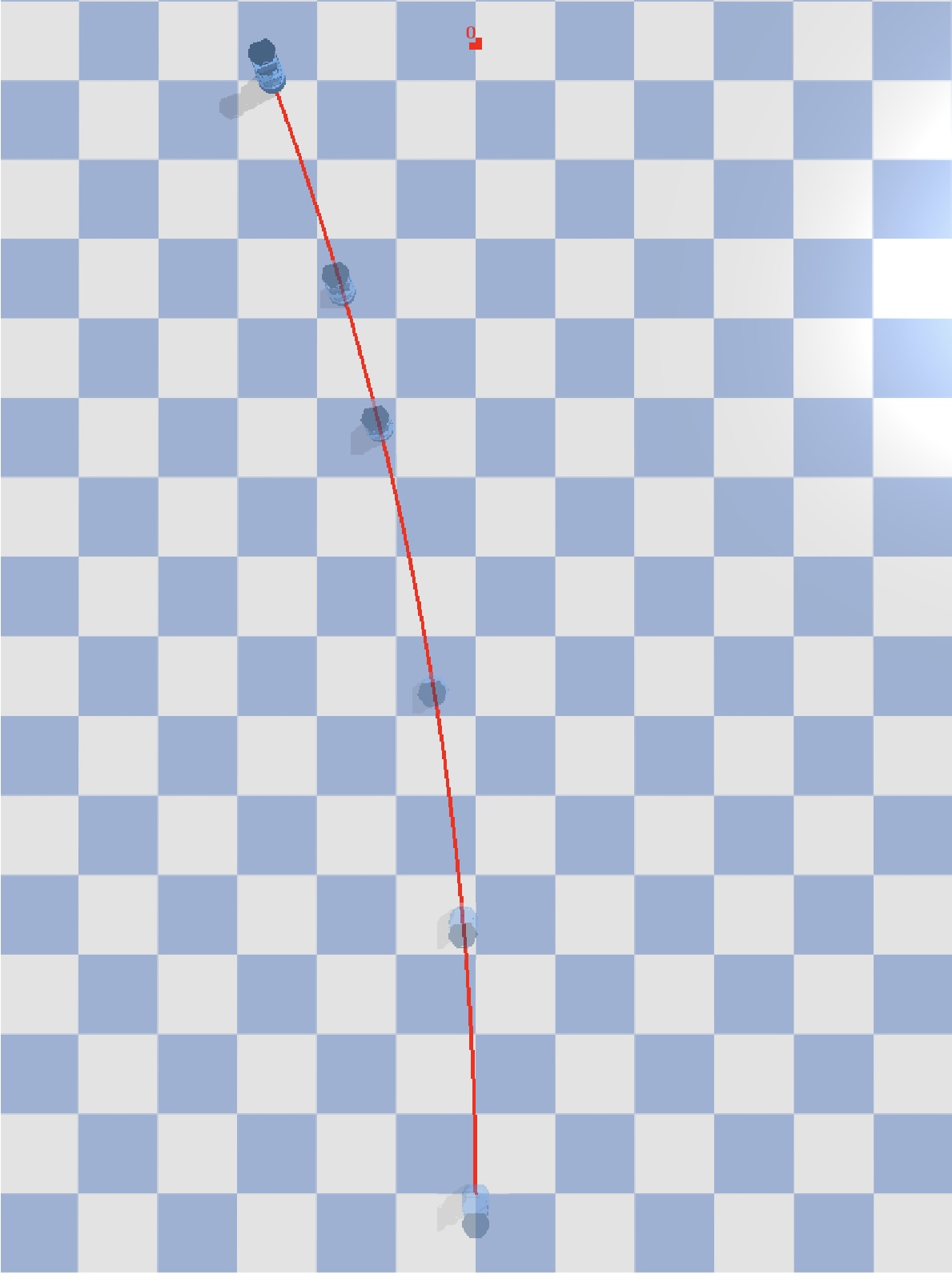}}
		\hfill  
		\subfigure[Real robot platform]{\includegraphics[height=0.17\textwidth]{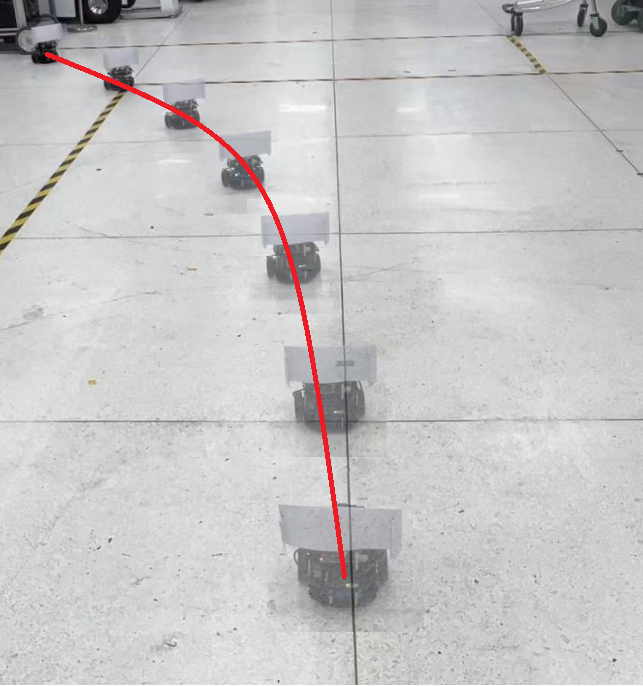}}
		\caption{Comparison between the simulators and the real world. Agents on all platforms execute commands with $v=1$ m/s, $\omega=0$ rad/s. The direct travel distance of the agent in the simulator is 10 m, and the distance of the real robot is 5 m.}
		\label{Simulator}
	\end{figure}

	A fundamental challenge in RL pertains to the significant sim-to-real gap observed when policies trained in conventional simulators are deployed on physical platforms. As shown in Fig.~\ref{Simulator} (a), (c), this discrepancy displays primarily as unrealistic trajectory generation due to oversimplified physical modeling ($x^{t+1}=x^{t}+v^{t}\Delta t+\frac{1}{2}a^t\Delta t^2$) within 2D simulation environments. Conventional simulators neglect essential forces and complex interactions inherent in real-world systems, producing idealized trajectories that diverge markedly from actual robot behaviors and inadequately capture real-world dynamics. Consequently, agents exhibit deficient robustness when confronted with real-world interference like spatiotemporal sparsity, emergent obstacles, and dynamic environmental changes. This deficiency extends to perceptual degradation stemming from deterministic sensor modeling. The absence of physically perturbations, insufficient exposure to critical failure states during training, and the lack of structured mechanisms to systematically induce and recover from near-failure states further compound these deficiencies. The large gap with the physical world limits policy generalization and obstructs the learning of robust fault recovery policies.
	
	To address these limitations comprehensively, a physics-enhanced simulation framework, PhysReplay-Simlab, leveraging PyBullet \cite{pybullet}, is developed. Fig.~\ref{Simulator} (b) shows the core innovation of PhysReplay-Simlab lies in modeling state transitions through $x^{t+1}=\mathcal{T}(x^t, v^t, a^t, f^t,...)$, where $\mathcal{T}$ represents the physics engine's transition function. This intrinsic generates realistic deviations and physical perturbations (e.g., simulated wheel slip), producing trajectories with significantly improved congruence with physical platforms compared to deterministic models. Furthermore, Gaussian noise injection during training specifically targets perceptual robustness by decoupling policy dependence on idealized sensor inputs.
	
\begin{figure}[htbp]  
		\centering  
		\includegraphics[width=0.45\textwidth]{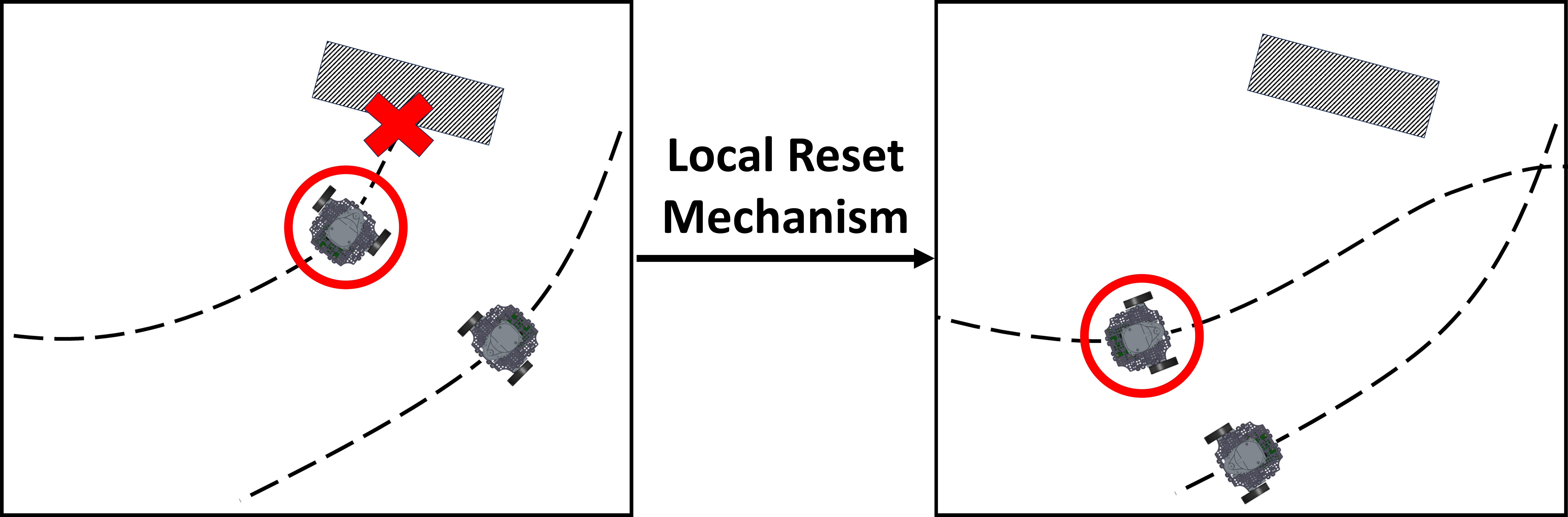}
		\caption{Illustration of the local replay mechanism. A near-collision state is highlighted (red circle). Upon collision, the mechanism resets the affected agent to its state N steps prior, while other agents continue their motion unaffected.}
		\label{local reset}
	\end{figure}

	PhysReplay-Simlab integrates a local replay mechanism to address challenges in dense multi-agent scenarios, including spatiotemporal sparsity, emergent obstacle prediction, and rapid adaptation to dynamic obstacles. As shown in Fig.~\ref{local reset}, upon collision, the agent is reset to its state from N=300 simulation steps prior, generating near-collision scenarios for training. The replay mechanism accelerates collision data accumulation through focused adversarial sampling. To mitigate training instability caused by repeated short-term collisions, agents exceeding a threshold collision are randomly reset to valid states within an iteration, promoting exploration diversity while maintaining trajectory continuity.
	
	PhysReplay-Simlab provides a scalable and adaptable platform for diverse applications. It integrates dynamic agents and random obstacles with adjustable control frequencies, enabling flexible adaptation to evolving training requirements. The transitions of the physics engine generate trajectories with realistic deviations, substantially enhancing behavioral alignment with physical platforms. The synergistic combination of physical perturbations and the local replay mechanism boosts policy resilience against environmental randomness and critical edge cases. Implemented within the Gym architecture \cite{Gym}, the framework ensures interface consistency and modular design, supporting functional reconfigurability. This adaptability is demonstrated by its straightforward modification for Safe RL evaluation \cite{SafeRL}, requiring only minimal changes to two core functions.
	
	\subsection{LSTP-Nav framework}
	
	\subsubsection{\textbf{LSTP-Net}}
	
	Lightweight Spatiotemporal Policy Network (LSTP-Net) employs a temporal module based on a Gated Recurrent Unit (GRU) to encode sequential LiDAR observations, mitigating the historical information loss inherent in the POMDP. The GRU is chosen for LSTP-Net because its gated mechanism (update and reset gates) selectively retains relevant historical states and forgets less critical information over time, effectively capturing long-range temporal dependencies. By processing stacked LiDAR frames, the GRU in LSTP-Net integrates crucial temporal information while maintaining computational efficiency compared to more complex recurrent architectures like LSTMs. This GRU-based policy demonstrates robust performance in cluttered environments and scenarios. However, LSTP-Net's effectiveness diminishes in high-density, dynamic multi-agent configurations due to the policy's over-reliance on historical patterns, which causes it to overlook immediate environmental threats.
	
	To address this limitation, LSTP-Net introduces an attention mechanism to dynamically balance LiDAR perception encoding. As illustrated in Fig.~\ref{policy2}, LiDAR observation sequences $o_s^t \in \mathbb{R}^{B \times 5 \times N_{\text{laser}}}$ are processed by LSTP-Net's two-layer GRU module (hidden size $d_h=256$), generating encoded state sequences $H \in \mathbb{R}^{B \times 5 \times 256}$:
	
	\begin{equation}
		H = \text{GRU}(o_s^t).
	\end{equation}
	
	The attention mechanism employs the GRU output of the last frame $H^{5}$ as the query vector $Q \in \mathbb{R}^{B \times 1 \times 256}$, while utilizing the full GRU output $H$ as key $K \in \mathbb{R}^{B \times T \times 256}$ and value $V \in \mathbb{R}^{B \times T \times 256}$. The context vector $C \in \mathbb{R}^{B \times 256}$ is computed via multi-head attention (4 heads):
	
	\begin{align}
		C &= \text{Concat}(h_1, \ldots, h_h) \cdot \mathbf{W}^O, \\  
		h_i &= \text{Attention}(Q\mathbf{W}_i^Q, K\mathbf{W}_i^K, V\mathbf{W}_i^V), \\  
		\text{Attention}&(Q, K, V) = \text{Softmax}(\frac{QK^T}{\sqrt{d_k}})V.
	\end{align}
	
	Concurrently, goal and velocity inputs $[o_g^t, o_v^t] \in \mathbb{R}^{B \times 4}$ undergo residual encoding:
	
	\begin{align}
		S_{\text{enc}} &= \mathbf{W}_{\text{res}}(\text{ELU}(\mathbf{W}_{\text{enc}}[o_g^t, o_v^t]) + \mathbf{W}_{\text{enc}}[o_g^t, o_v^t]).
	\end{align}
	
	The final feature representation combines context and state through concatenation $\text{data} = [C,\ S_{\text{enc}}] \in \mathbb{R}^{B \times 512}$. The actor network generates motion policies $[v^t, \omega^t]$ via:
	
	\begin{align}
		\mu &= \mathbf{W}_{a3} \cdot \text{ELU}(\mathbf{W}_{a2} \cdot \text{ELU}(\mathbf{W}_{a1} \cdot \text{data})), \\
		\sigma &= \text{trainable parameter} \in \mathbb{R}^2.
	\end{align}
	
	The critic network estimates state value through:
	
	\begin{equation}
		V_\phi(o^t) = \mathbf{W}_{c3} \cdot \text{ELU}(\mathbf{W}_{c2} \cdot \text{ELU}(\mathbf{W}_{c1} \cdot \text{data})).
	\end{equation}
	
	All intermediate layers ($\mathbf{W}_{enc}, \mathbf{W}_{res} \in \mathbb{R}^{256\times256}$ and $\mathbb{R}^{256\times256}$, $\mathbb\mathbf{W}_{a1/a2}, \mathbf{W}_{c1/c2} \in \mathbb{R}^{512\times256}$ and $\mathbb{R}^{256\times128}$) utilize ELU activations, while output layers ($\mathbf{W}_{a3} \in \mathbb{R}^{128\times2}$, $\mathbf{W}_{c3} \in \mathbb{R}^{128\times1}$) operate linearly. This design enables adaptive focus on critical LiDAR features while preserving goal-velocity information via residual connections, enhancing robustness against perceptual uncertainties.
	
	This lightweight network (1.15 M parameters) achieves $> 300$ Frames Per Second (FPS) real-time navigation on edge devices by employing streamlined GRUs (replacing complex LSTMs) for temporal processing. It maintains accuracy in dynamic scenarios by balancing GRU-based historical modeling with attention-driven threat response, dynamically focusing on critical LiDAR features. Efficient input integration via residual connections and feature compression into unified representations minimizes computational overhead.
	
	\subsubsection{\textbf{Training Strategy}}
	
	The LSTP-Nav training strategy is described as Algorithm \ref{PPOMA}. In line 1, the policy network \(\pi_{\theta}\), the value function network \(V_{\phi}\), and hyperparameters are initialized. Lines 2-9 outline the data collection phase: for each iteration (up to \(M\)), the agents interact with the environment over \(T_{\text{max}}\) timesteps. Specifically, for each robot \(i = 1, 2, \ldots, N\), the policy network \(\pi_{\theta}\) generates actions \(a_i^t\) based on observations \(o_i^t\), collecting rewards \(r_i^t\), while the value network \(V_{\phi}\) estimates state values \(V_{\phi}(o_i^t)\).  
	
	In lines 10-15, the trajectories undergo post-processing: for each robot \(i\) and timestep \(t\) (processed in reverse order from \(T_{\text{max}}\) to 1), the algorithm computes the temporal difference (TD) error \(\delta_i^t\) as:
	\begin{equation}
		\delta^t = r^t + \gamma V_{\phi}(o^{t+1}) - V_{\phi}(o^t),
		\label{td}
	\end{equation}
	and subsequently calculates the Generalized Advantage Estimate (GAE) \cite{GAE} \(A^t\) using:
	\begin{equation}
		A^t = \sum_{k=t}^{T_{\text{max}}} (\gamma \lambda)^{k-t} \delta^k.
		\label{GAE}
	\end{equation}
	The return \(V^t\) is then derived as \(V^t = A^t + V_{\phi}(o^t)\). 
	
	In line 16, the collected trajectories are merged and shuffled to form training batches. Line 17 saves the current policy and value network parameters (\(\theta_{\text{old}} = \theta\), \(\phi_{\text{old}} = \phi\)) before optimization. Lines 18-22 describe the parameter update phase: for \(K\) epochs, the policy loss \(L^P\), value loss \(L^V\), and entropy loss \(L^E\) are computed by: 
	\begin{align}
		L^{P} &= -\min\Bigg( \frac{\pi_{\theta}(a|s)}{\pi_{\theta_{\text{old}}}(a|s)} A, \nonumber \\
		&\qquad\quad \text{clip}\left(\frac{\pi_{\theta}(a|s)}{\pi_{\theta_{\text{old}}}(a|s)}, 1 - \epsilon, 1 + \epsilon\right) A \Bigg) \label{eq:policy_loss}, \\
		L^{V} &= \max\Bigg( (V_{\theta} - V_{\text{targ}})^{2}, \nonumber \\
		&\qquad\quad \left( \text{clip}\left(V_{\theta}, V_{\theta_{\text{old}}} - \varepsilon, V_{\theta_{\text{old}}} + \varepsilon\right) - V_{\text{targ}} \right)^{2} \Bigg) \label{eq:value_loss}, \\
		L^{E} &= H\left( \pi_\theta (a^t|o^t) \right), \label{eq:entropy_loss}
	\end{align}
	where $V_{\text{targ}}=A+V_\phi(s)$. The composite loss function is defined as $L^{{Total}}=L^{{P}}+\alpha L^{V}+\beta L^{E}$, where $\alpha$ and $\beta$ are empirically tuned hyperparameters. Network parameters are optimized using the Adam optimizer \cite{Adam} with respect to $L^{Total}$. Finally, the algorithm terminates upon reaching the maximum iteration count in Line 23.
	
	\begin{algorithm}
		\caption{LSTP-Nav Training Strategy}
		\begin{algorithmic}[1] 
			\State Initialize policy network $\pi_{\theta}$, value function network $V_{\phi}$, and set hyper-parameters
			\For{$iteration = 1, 2, \ldots$, $M$}
			\For{$t = 1, 2, \ldots, T_{\text{max}}$}
			\For{each robot $i = 1, 2, \ldots, N$}
			\State Run policy network $\pi_{\theta}$ to collect $\{o_i^t, a_i^t, r_i^t\}$
			\State Run value function network to collect $\{V_{\phi}(o_i^t)\}$
			\EndFor
			\EndFor
			\For{each robot $i = 1, 2, \ldots, N$}
			\For{$t = T_{\text{max}}, T_{\text{max}} - 1, \ldots, 1$}
			\State Calculate TD-error using (\ref{td})
			\State Estimate advantages using (\ref{GAE})
			\State Calculate Return $V_i^t = A_i^t + V_{\phi}(o_i^t)$
			\EndFor
			\EndFor
			\State Merge and shuffle the collected trajectories
			\State $\theta_{\text{old}} = \theta$, $\phi_{\text{old}} = \phi$
			\For{$k$ in $K$}
			\State Calculate loss components using (\ref{eq:policy_loss}), (\ref{eq:value_loss}), (\ref{eq:entropy_loss})
			\State Calculate total loss: $L^{\text{Total}} = L^{P} + \alpha L^V + \beta L^{E}$
			\State Update the parameters of $\pi_{\theta}$ and $V_{\phi}$ 
			\EndFor
			\EndFor
		\end{algorithmic}
		\label{PPOMA}
	\end{algorithm}
	
	\textit{Reward Function:} Conventional RL navigation employs dense rewards combining goal, obstacle, and auxiliary rewards (e.g., RVO-based constraints or control limits). However, excessive reward components risk training instability and complex hyperparameter tuning. To eliminate volatile constraints shown to hinder convergence and limit final policy performance, this work streamlines the reward function exclusively to goal and obstacle rewards, which is:
	\begin{align}
		R &= r_g + r_c,
	\end{align}
	where $r_g$ represents the goal reward, $r_c$ represents the obstacle reward. 
	
	Specifically, $r_g$ incentivizes trajectory efficiency by measuring the reduction in Euclidean distance to the goal between timesteps, formally defined as:
	\begin{align}
		r_g &= \begin{cases} 
			r_{\text{arrival}} & \text{if } \| \mathbf{p}_i^t - \mathbf{g}_i \| < 0.1 \\
			w_g \left( \| \mathbf{p}_i^{t-1} - \mathbf{g}_i \| - \| \mathbf{p}_i^t - \mathbf{g}_i \| \right) & \text{otherwise}
		\end{cases},
	\end{align}
	where $w_g$ are the weight of $r_g$ and $d_i^t$ represents the minimum distance between agent $i$ and obstacles at timestep $t$. The goal reward provides dense incremental rewards for a progressive goal approach while reserving sparse, high-value rewards for final success.
	
	\begin{figure}[htbp]  
		\centering  
		{\includegraphics[width=0.45\textwidth]{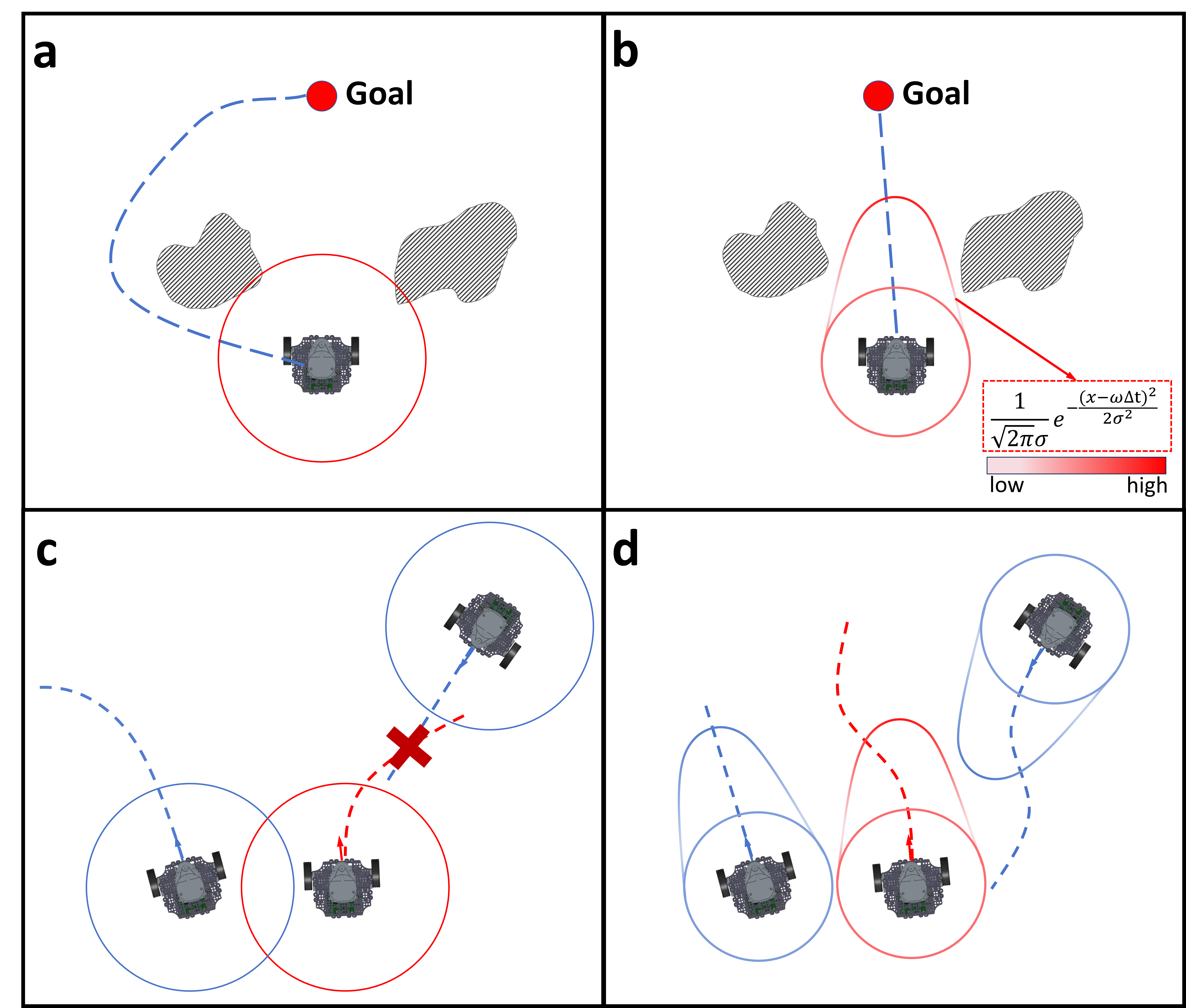}}
		\caption{Comparison between conventional reward functions and our solution in narrow passages and multi-agent navigation. (a) and (c) reveal suboptimal actions caused by $r_c$: (a) an agent circumnavigating obstacles, and (c) hazardous multi-agent interactions. In contrast, (b) and (d) showcase our approach's improvements: (b) successful narrow gap traversal through balanced risk evaluation, and (d) safe multi-agent coordination via comprehensive long-term hazard anticipation.}  
		\label{classic reward function}
	\end{figure} 
	While $r_g$ incentivizes goal advancement, the obstacle reward $r_c$ enforces safety via collision avoidance. However, conventional $r_c$ formulations exhibit critical limitations that compromise navigation efficiency.	As Fig. \ref{classic reward function} illustrates, the classic obstacle reward discourages proximity to obstacles even when traversing a narrow gap is optimal, frequently inducing unnecessary detours. This leads to suboptimal circumnavigation behaviors in single-agent scenarios. In multi-agent settings, the reward only considers imminent threats, neglecting distant agents posing potential long-term risks. Most critically, traditional $r_c$ formulations lack actionable policy gradients for optimal action selection during critical encounters. To resolve these deficiencies, this work proposes a novel heading stability reward (HS reward):
	\begin{align}
			r_c &= \begin{cases} 
				r_ { collision } & \text {if } d_i^t < 0.01 \\
				w_{c}\left(z_{max}-z_{i}^t\right) & \text { otherwise}
			\end{cases},
	\end{align}
	where $w_c$ are the weight of $r_c$, $z_{max}$ denotes the maximum detection range of LiDAR, and $z_i^t$ represents the LiDAR observations for agent $i$ at timestep $t$. The weight vector $w_c$ follows a Gaussian distribution centered at the current angular displacement $\omega\Delta t$: 
	\begin{equation}
		\begin{array}{l}
			w_c = \frac{1}{\sqrt{2\pi}\sigma}e^{-\frac{(x-\omega\Delta t)^2}{2\sigma^2}},
		\end{array}
	\end{equation}
	where $\omega$ is the angular velocity, $\sigma$ controls the spread of the distribution, $\Delta t$ represents the control time interval, and $x$ denotes the angular position relative to $\omega\Delta t$. The weights decrease symmetrically from the center $\omega\Delta t$ according to the Gaussian profile, as illustrated in Fig. \ref{classic reward function} (b). 
	
	The HS reward integrates the agent's angular velocity to obstacle avoidance, which accomplishes three main goals: prioritizes obstacles in the predicted heading direction, maintains awareness of all proximal obstacles, and encourages deliberate angular velocity selection. The HS reward improves policy performance in challenging scenarios by promoting deliberate angular adjustments, providing clearer control gradients, and balancing directional focus with comprehensive obstacle awareness.
	
	\begin{figure}[htbp]  
		\centering  
		\subfigure[Without the HS reward]{\includegraphics[width=0.45\textwidth]{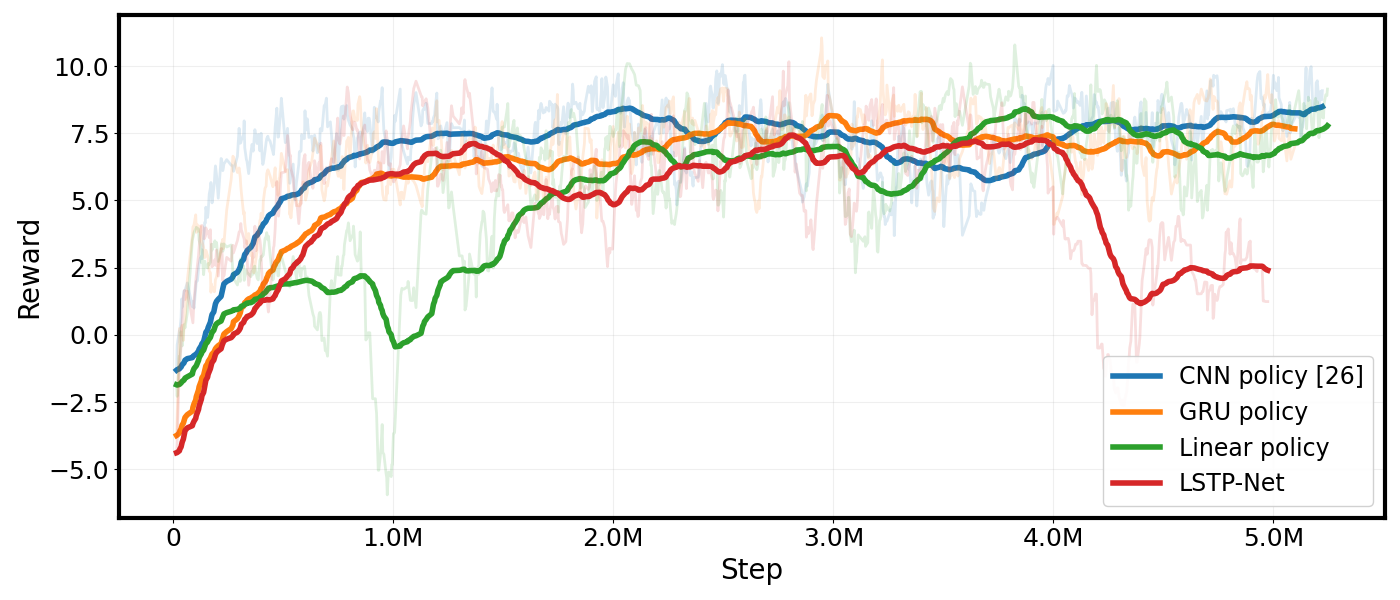}}\\
		\vfill 
		\subfigure[With the HS reward]{\includegraphics[width=0.45\textwidth]{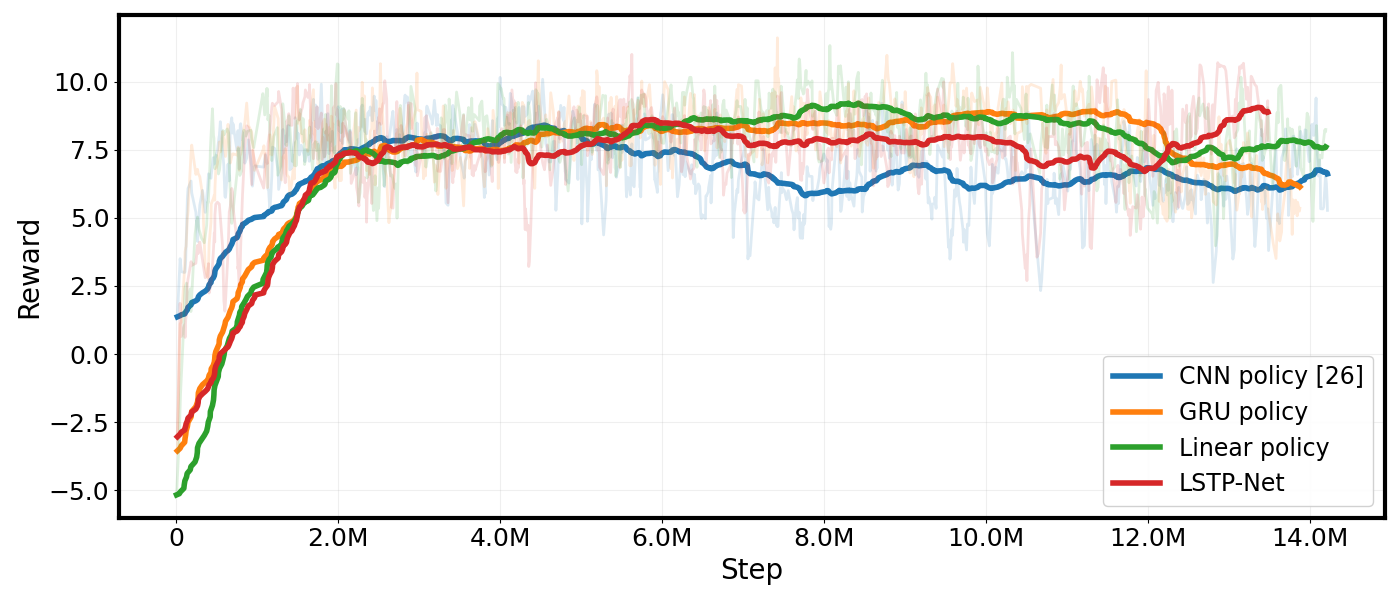}}
		\caption{Contrasting training dynamics and convergence between (a) conventional reward (instability, avoidance) and (b) the HS reward (rapid convergence, sustained near-optimal behavior).}
		\label{reward training compare}
	\end{figure}  
	
	Fig.~\ref{reward training compare} shows the reward convergence during training. The policy denoted in the legend is explained in Section~\ref{experiments}. Conventional reward formulations induce training instability: certain policies fail to maintain optimal actions as training progresses, while others struggle to discover effective behavioral strategies. Non-convergent policies exhibit avoidance behaviors (e.g., freezing when facing obstacles). In contrast, the HS reward significantly stabilizes training dynamics, enabling all policies to achieve rapid convergence and sustain near-optimal behavior across extended training epochs. This enhancement directly translates to the robust navigation performance observed in our simulation trials. 
	
\begin{figure}[htbp]  
		\centering  
		\subfigure[Single-agent scenario]{\includegraphics[height=0.225\textwidth]{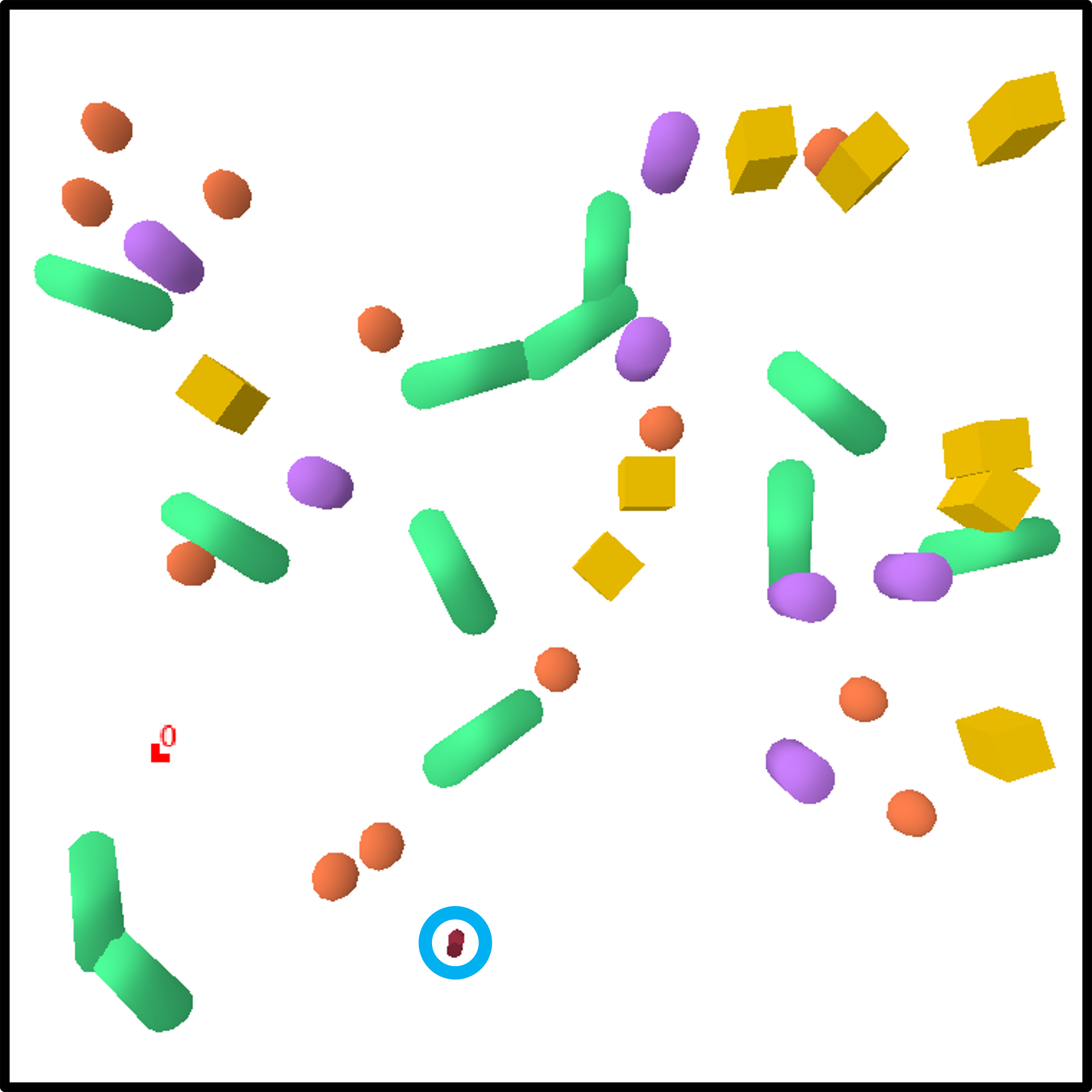}}
		\hfill 
		\subfigure[Multi-agent scenario]{\includegraphics[height=0.225\textwidth]{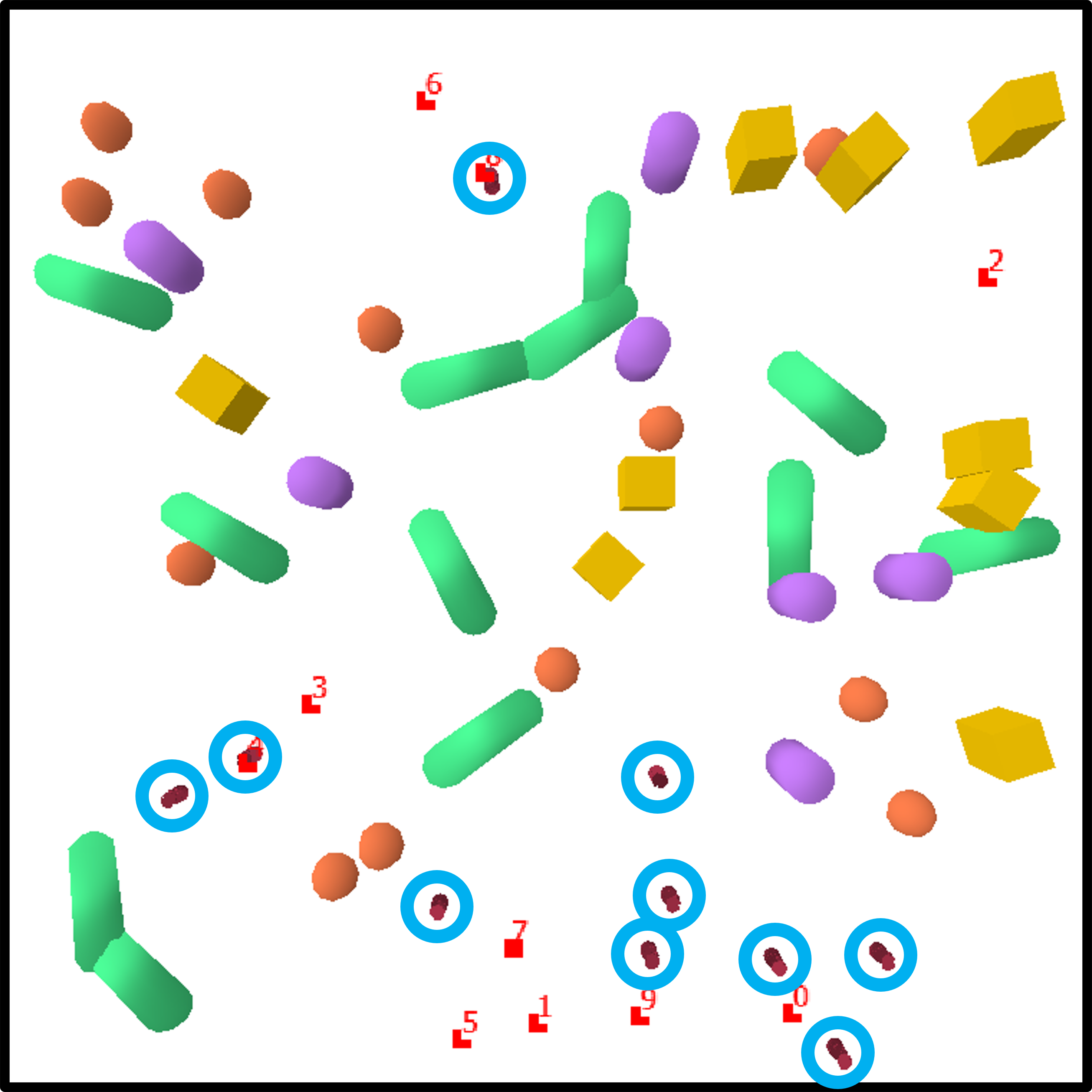}}
		\caption{The two scenarios during training: (a) single-agent scenario: 30 obstacles are randomly placed in an area of 8 m$\times$ 8 m. (b) multi-agent scenario: 10 agents and 35 obstacles are randomly placed within an area of $\text{10 m}\times \text{10 m}$, where the blue circles circle the agents; yellow squares, purple cylinders, and orange balls represent random obstacles; red square represents the agent's navigation target point; obstacles, the initial position of the robot, and the navigation target point are all randomly generated. }
		\label{training scenarios}
	\end{figure}

	\textit{Training Scenarios:} The training procedure employs two distinct environmental scenarios: single-agent and multi-agent, which are shown in Fig. \ref{training scenarios}. Each scenario contains randomized rigid obstacles: spheres (0.5 m radius), cubes ($1\times 1\times 1$ m$^3$), capsules (2 m length, 0.5 m radius), and cylinders (1 m length, 0.5 m radius). Obstacle positions are uniformly sampled within a specified area with unconstrained orientations ($\theta \in [0, pi)$ rad), creating an unstructured navigation challenge. The agents are also randomly placed in the scenarios. The training environment is dynamically generated in real-time without providing agents with any prior knowledge. 
	
	The training pipeline implements a curriculum learning paradigm by progressively scaling environmental complexity. Single-agent training first validates the policy in an environment containing 5 random obstacles, establishing core collision avoidance skills. Mastery of this task triggers advancement to a high-difficulty setting with 30 random obstacles, where the policy refines navigation in dense clutter. For multi-agent scenarios, the curriculum shifts to cooperative adaptation: 10 agents are initialized randomly amidst 35 random obstacles, testing zero-shot transfer from single-agent training to multi-agent navigation. By progressively evolving obstacle density (sparse to dense) and agent scale (solo to multi-agent), this structured approach mimics skill acquisition. Such tiered training mitigates premature convergence in complex scenarios while enhancing policy generalization. A 2500-step episode limit for adequate navigation time, and these combined design choices ensure consistent learning dynamics across diverse scenarios.
	
	\section{EXPERIMENT EVALUATION}
	\label{experiments}
	\subsection{Experimental Setup}
	All simulation experiments are conducted on a computational server equipped with an NVIDIA GeForce RTX 4090 GPU and an Intel Xeon 8336C processor, with models implemented using the PyTorch framework under optimized parameter configurations. For real-world deployment, experiments are conducted on a basic TurtleBot3 (TB3) platform (specifications detailed in Table \ref{tab3}). To meet the experimental requirements, the platform's LiDAR system is configured to operate within a reduced $0.8\pi$ field of view (FOV), utilizing 130 beams with interval sampling. Environmental perception is achieved by combining LiDAR and wheel odometry, while maintaining policy parameters consistent with the simulation setup. Notably, the system achieves a computation frequency of over 40Hz during real-world deployment on the CPU-based platform.
	
	\begin{table}[htbp]  
		\centering  
		\caption{TB3 Basic Platform Specifications}  
		\label{tab3}  
		\renewcommand{\arraystretch}{1.1} 
		\begin{tabularx}{\linewidth}{|l|X|}  
			\hline  
			\textbf{Category} & \textbf{Specification} \\ \hline  
			Physical Dimensions & 138 mm × 178 mm × 192 mm (L×W×H) \\ \hline  
			Drive System & Dynamixel XL430-W250 \\ \hline  
			Control Frequency & 10 Hz \\ \hline
			Motion Capabilities &  
			Max linear velocity: 0.22 m/s, Max angular velocity: 2.84 rad/s \\ \hline  
			Sensing &  
			LiDAR (360 Laser Distance Sensor LDS-01) 0.012-0.35 m range, ±15 mm accuracy, 1° resolution, $2\pi$ rad FOV; 
			IMU: Gyroscope 3 Axis; 
			Wheel odometry: ±0.05 m positional accuracy \\ \hline  
			Computation & Raspberry Pi 3 Model B \\ \hline  
			Experimental Config &  
			130 LiDAR beams ($0.8\pi$ FOV), policy output scaled to platform limits \\ \hline  
		\end{tabularx}
	\end{table}
	
	The proposed method (LSTP-Net) is evaluated through comparative experiments in both simulated and real-world environments, with the following baseline methods: NH-ORCA \cite{ORCA-NH} and the CNN policy \cite{IJRR} with the conventional reward function. Ablation studies are performed to assess the contributions of key architectural components.
	
	The evaluation metrics include: (1) Success Rate (SR): the rate of which the agents reach the goal within time constraints without collisions; (2) Collision Rate (CR): the agents collide with each other and obstacles; (3) Trap Rate (TR): occurrences of timeout without goal achievement or collisions; (4) Average Steps (AS): average simulation steps required for successful navigation. Simulation steps are adopted as the temporal metric instead of the cost time to eliminate computational platform dependencies that could distort real-world performance assessment. All metrics are computed over 1000 independent trial runs to ensure statistical significance.
	
	\subsection{Simulation Experiments and Results} 
	To comprehensively evaluate the LSTP-Net (ours), we conduct comparative experiments against established baselines: \textbf{NH-ORCA} \cite{ORCA-NH} and \textbf{CNN policy} \cite{IJRR} with conventional rewards in simulated environments. Additionally, ablation studies are performed to isolate the contribution of key architectural components: (1) \textbf{removing the attention mechanism (GRU policy)}, and (2) \textbf{replacing the GRU with linear layers (Linear policy)}.
	
	\begin{table}[htbp]  
		\centering  
		\caption{Experiment with 1 agent and 30 obstacles}  
		\label{tab1}  
		\renewcommand{\arraystretch}{1.1} 
		\begin{tabular}{|c|c|c|c|c|c|}  
			\hline  
			& SR (\%) $\uparrow$ & CR (\%) $\downarrow$ & TR (\%) $\downarrow$ & AS $\downarrow$ \\ \hline 
			NH-ORCA \cite{ORCA-NH} & 79.20 & 17.80 & 3.00 & 342.49\\ \hline
			CNN policy \cite{IJRR} & 96.60 & 0.90 & 2.50 & \textbf{337.05} \\ \hline   
			GRU policy & 95.7 & 3.1 & 1.2 & 420.65 \\ \hline  
			Linear policy & \textbf{99.20} & \textbf{0.70} & \textbf{0.10} & 355.68 \\ \hline  
			\textbf{Ours} & 98.20 & 1.70 & \textbf{0.10} & 403.10 \\ \hline
		\end{tabular}  
	\end{table} 	
	
	\begin{table}[htbp]  
		\centering  
		\caption{Experiment with 10 agents and 35 obstacles}  
		\label{tab2}  
		\renewcommand{\arraystretch}{1.1} 
		\begin{tabular}{|c|c|c|c|c|c|}  
			\hline  
			& SR (\%) $\uparrow$ & CR (\%) $\downarrow$ & TR (\%) $\downarrow$ & AS $\downarrow$ \\ \hline 
			NH-ORCA \cite{ORCA-NH} & 77.26 & 20.44 & 2.30 & 499.78 \\ \hline  
			CNN policy \cite{IJRR} & 83.39 & 14.93 & 1.68 & 486.49 \\ \hline  
			GRU policy & 87.10 & 9.26 & 3.64 & 640.38 \\ \hline  
			Linear policy & 86.13 & 13.86 & \textbf{0.01} & \textbf{463.20} \\ \hline
			\textbf{Ours} & \textbf{92.97} & \textbf{2.63} & 4.40 & 822.77  \\ \hline
		\end{tabular}  
	\end{table} 
	
	The single-agent test environment features a robot navigating among 30 obstacles within a $8\,\text{m} \times 8\,\text{m}$ area, with results in Table~\ref{tab1}. Compared to the CNN policy, the LSTP-Net achieves a $1.6\%$ higher SR and $2.4\%$ lower TR but exhibits a $0.8\%$ higher CR. The CNN policy's $2.5\%$ TR predominantly originates from navigation failures in narrow passages, a limitation attributable to its conventional reward function design.	Notably, the Linear policy demonstrates optimal performance in SR ($99.20\%$), CR ($0.7\%$), and TR ($0.1\%$), while the CNN policy achieves the fastest navigation (337.05 AS). In contrast, NH-ORCA's suboptimal performance ($79.20\%$ SR, $17.80\%$ CR) stems from fundamental challenges in obstacle modeling, particularly in dense configurations where inadequate geometric modeling induces recurrent entrapment.
	
	The multi-agent navigation environment is configured with 10 robots and 35 obstacles in a $10\,\text{m} \times 10\,\text{m}$ area. The results shown in Table~\ref{tab2} demonstrate that the LSTP-Net achieves 9.58\% higher SR than the CNN baseline while reducing CR by 12.30\%. Notably, the Linear policy attains the lowest TR (0.01\%) and fastest navigation speed (463.20 AS), outperforming NH-ORCA by 2.29\% in TR reduction. The CNN policy maintains the navigation efficiency (486.49 AS), though its 1.68\% TR primarily stems from the same reason as the single-agent scenario. Comparative analysis reveals NH-ORCA's fundamental limitations in collision avoidance (20.44\% CR) and local navigation (77.26\% SR)  due to fixed interaction rules and inadequate obstacle modeling.
	
	\begin{figure*}[htbp]  
		\centering  
		\includegraphics[width=0.9\textwidth]{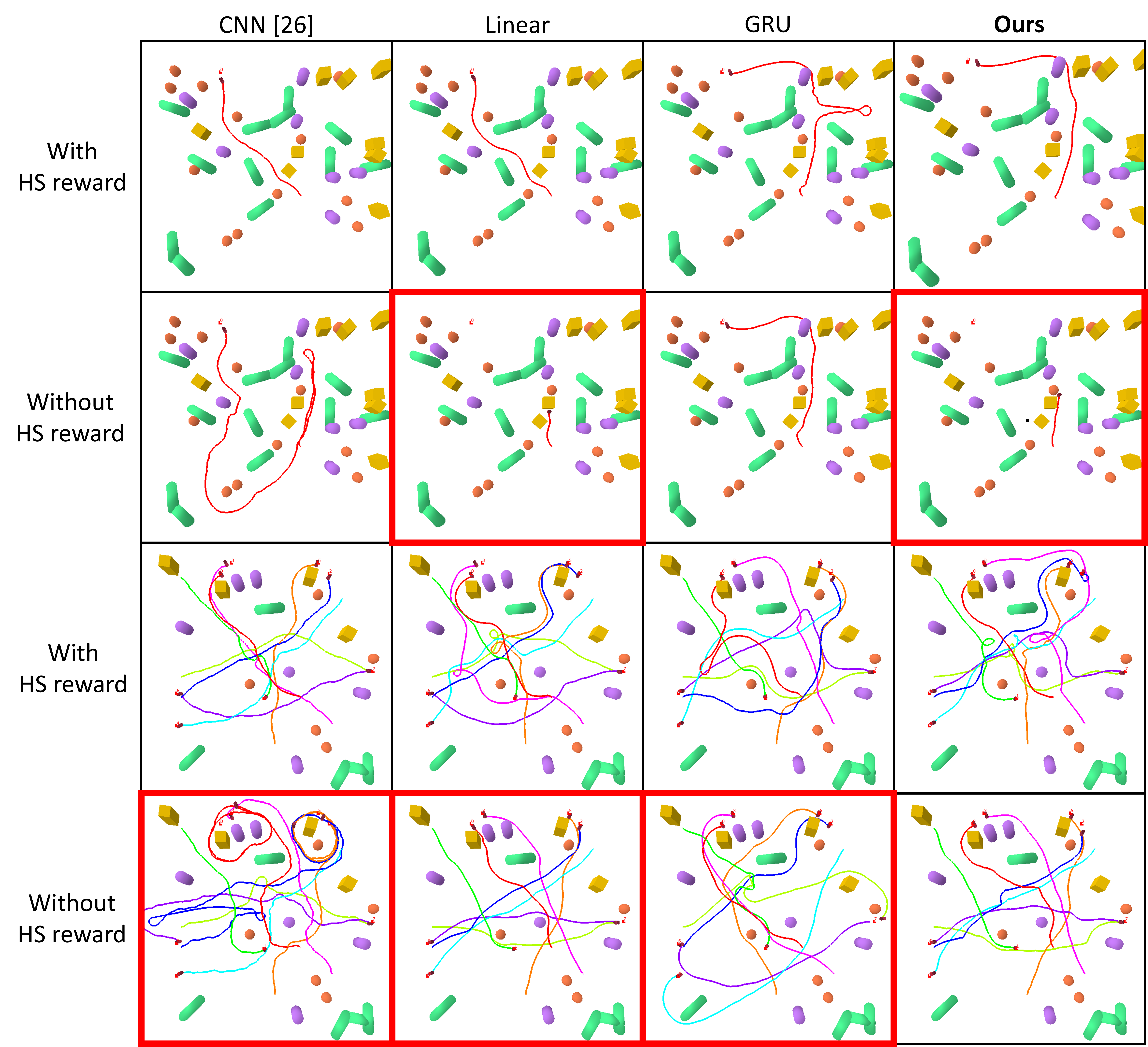}  
		\caption{Policies' performance are evaluated across test scenarios as shown in the figure. \textbf{Red frames indicate failure cases: agents missing goals or colliding with obstacles/other agents.} Columns represent different policies from left to right: CNN, Linear, GRU, and ours (LSTP-Net) policies. Rows correspond to experimental conditions: rows 1-2 display results for the single-agent scenario ($\pm$ HS reward), while rows 3-4 present the multi-agent scenario ($\pm$ HS reward). The sequential trajectory colors correspond as follows: 0: red, 1: green, 2: blue, 3: magenta, 4: cyan, 5: orange, 6: purple, 7: lime green.}  
		\label{simulation test}
	\end{figure*}
	
	Fig.~\ref{simulation test} demonstrates the performance of policies trained with or without the HS reward. Red frames highlight failure cases where agents either failed to reach their goals or collided with obstacles or other agents. The test scenarios specifically evaluate the policies' ability to navigate narrow passages and manage multi-agent interactions. The first two rows depict single-agent navigation within a 10 m $\times$ 10 m area containing 40 obstacles. Narrow passages are intentionally formed by these obstacles for the agent to traverse successfully. The last two rows present multi-agent interaction within a 10 m $\times$ 10 m area containing 20 obstacles and 8 agents. Agents are forced to interact closely in the central region while avoiding collisions with obstacles and navigating diverse encounter scenarios, specifically crossing, parallel movement, and head-on situations with other agents. The sequential trajectory colors correspond as follows: 0: red, 1: green, 2: blue, 3: magenta, 4: cyan, 5: orange, 6: purple, 7: lime green.
	
	In row 1, both the CNN and Linear policies successfully navigate the passage directly ahead without trajectory adjustments. Conversely, the LSTP-Net diverts to the right, traversing a passage on the opposite side of the scenario. Meanwhile, the GRU policy adopts a circuitous route, ultimately passing through a passage after significant detours.
	
	In row 2 (without HS reward), both LSTP-Net and Linear policies fail to reach the goal. These policies exhibit cautious but inefficient navigation: slow movement results from collision avoidance. The CNN policy attempts obstacle traversal but fails at narrow passages, resorting to longer alternative routes. The GRU policy maintains direct passage traversal capability like its HS reward counterpart, but demonstrates new collision vulnerability absent from HS rewards. This occurs because distal obstacles perturb trajectories during passage crossing, frequently triggering sharp left turns that increase collision risk with left-side obstacles.
	
	In row 3, which examines multi-agent interactions, the LSTP-Net and Linear policies demonstrate effective cooperative behaviors. Agents exhibit abilities such as waiting for others to pass and maneuvering around congested areas to avoid collisions; specific examples include Agents 1, 3, 4, and 7 (ours) and Agents 3, 5, and 7 (Linear). Notably, Agent 2 (LSTP-Net) efficiently approaches its goal by rapidly turning and backing towards it upon nearing the target after initially missing the goal. The CNN policy primarily learns collision avoidance through executing sharp turns, as demonstrated by Agents 0 and 1. In contrast, the GRU policy struggles in parallel movement scenarios, where agents (e.g., Agents 0 and 6) are frequently displaced from their intended course by neighboring agents. However, these agents typically manage to correct their trajectory afterward.
	
	In row 4 (without HS reward), only the LSTP-Net achieves smooth, successful navigation for all agents. Agents using this policy also demonstrate effective sharp turns, exemplified by Agents 0 and 2. The CNN policy successfully avoids collisions but executes discontinuous actions, leading to twisted and extended trajectories. Crucially, it fails to reach goals near obstacles. Agents 0, 2, and 5 oscillate around the obstacle adjacent to the goal instead of proceeding to the goal. The Linear policy generally avoids collisions with other agents. However, Agent 0 collides with an obstacle while approaching its goal. Although the trajectory visually passes through the obstacle, this is caused by the agent squeezing through the narrow gap between the obstacles. In contrast, the GRU policy fails to reach goals without collisions. Agents 1 and 7 collide, resulting in discontinuous trajectories, while the remaining agents navigate successfully.
	
	Experimental data confirms that the Linear policy achieves optimal static navigation (99.20\% SR, 0.70\% CR, 0.10\% TR in Table~\ref{tab1}), validating attention mechanisms' spatial perception strength in cluttered environments. However, its inferior multi-agent performance (86.13\% SR compared with LSTP-Net's 92.97\% SR in Table~\ref{tab2}) exposes dynamic environment limitations where GRU-based policies demonstrate superior dynamic adaptability. The LSTP-Net architecture successfully integrates attention with GRU modules to overcome these shortcomings, achieving the highest multi-agent success rate. Crucially, the HS reward significantly enhances robustness across all policies. When the HS reward is removed, LSTP-Net and Linear policies exhibit inefficient exploration (Fig.~\ref{simulation test} row 2), while the CNN policy frequently fails during passage traversal, resorting to excessively long paths. The GRU policy develops collision sensitivity despite maintaining trajectory consistency, evidenced by erratic turns near obstacles (Fig.~\ref{simulation test} row 4). Conversely, CNN policy produces discontinuous trajectories, Linear policy struggles at obstacle proximity, and GRU policy suffers persistent collisions without orientation guidance from HS.
	
	\subsection{Hardware Deployment} 
	In real-world deployment, due to its lightweight design, this policy can achieve computation frequency exceeding 40Hz on CPU-based platforms.
	
\begin{figure}[htbp]  
		\centering  
		\includegraphics[width=0.45\textwidth]{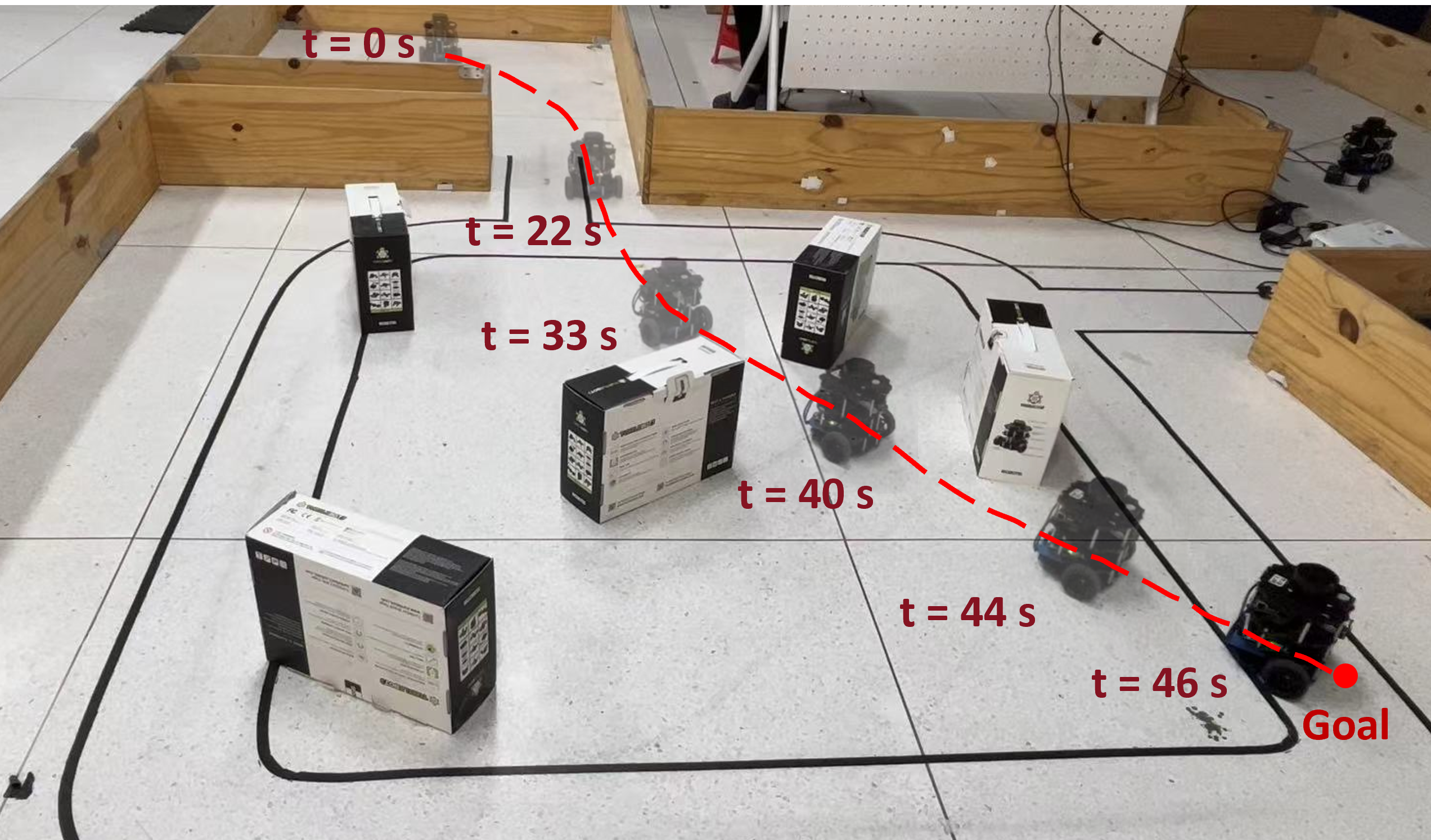}  
		\caption{Real-world single-agent experiment. The TB3 first passes through narrow passages, crosses through the obstacles, and finally reaches the goal point. The TB3 reaches the goal point in 46 s.}  
		\label{rse}
	\end{figure}

	In the single-agent scenario, the policy's performance is evaluated by controlling the TB3 robot through narrow passages. The experimental results demonstrate that the TB3 effectively navigates through constrained pathways while dynamically avoiding obstacles of unknown positions and geometries to reach the designated target location. As shown in Fig. \ref{rse}, the experiment begins with the TB3 positioned in a starting room. The robot successfully traverses a narrow passage, entering the target room within 22 seconds. Subsequently, it performs obstacle avoidance maneuvers, reaching the final goal point in 46 seconds. The navigation task is completed once the TB3 arrives at the predefined goal. The results confirm the policy's capability to enable autonomous navigation through narrow environments.
	
\begin{figure*}[htbp]  
		\centering  
		\includegraphics[width=\textwidth]{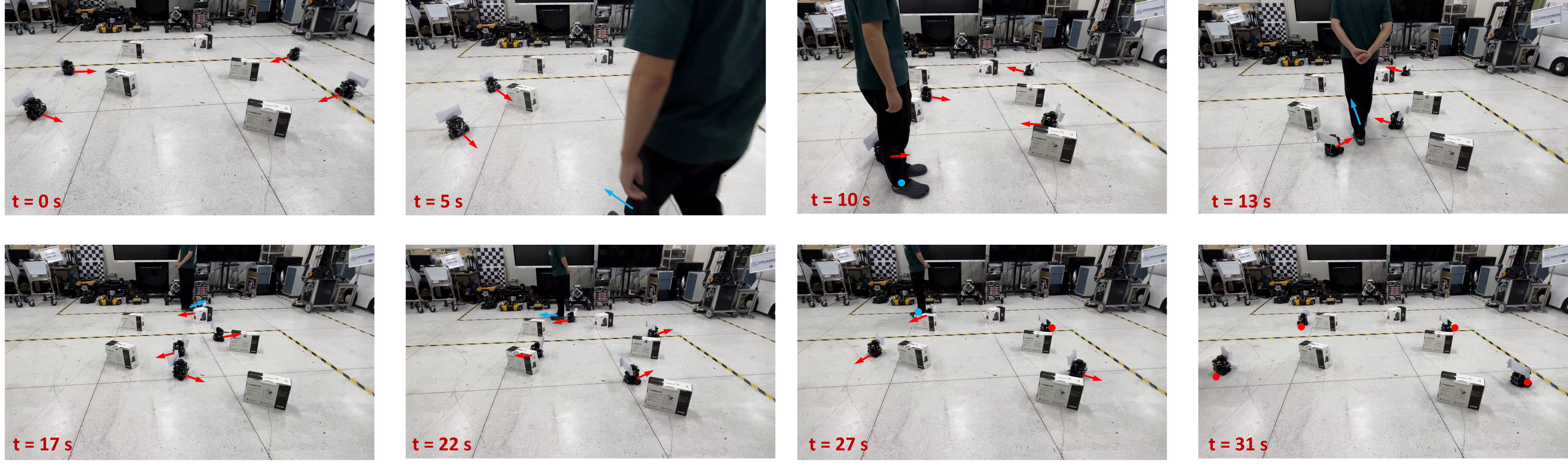}  
		\caption{Real-world multi-agent experiment. The red pattern indicates the state of TB3, the blue pattern indicates the state of a human, the arrow represents the direction of velocity, and the circle represents reaching the goal point (TB3) or remaining stationary (human). At t = 0 s, each TB3 navigates to the goal point. A human appears in the environment at t = 5 s, starting to interfere with TB3. The human first forces a TB3 to change its trajectory and crosses in front of two TB3s at t = 10 s and t = 13 s. Then, the human interferes with another TB3 again until it reaches the goal point for the rest of the time. All TB3s reach the goal points in 31 s.}  
		\label{rme}
	\end{figure*}

	In the multi-agent scenario, the policy's performance is assessed based on collision avoidance capabilities and responsiveness to dynamic disturbances. Each TB3 must navigate through a dense obstacle environment while avoiding interactions with other robots and human agents. For this evaluation, a human introduces dynamic disturbances by deliberately obstructing the TB3's paths. As depicted in Fig. \ref{rme}, the human first intercepts one TB3, prompting an immediate trajectory adjustment. The human then maneuvers between two TB3s, further testing their reactive navigation. The interference persists until the TB3s reach their respective goal points. Despite these disruptions, all TB3s successfully maintain collision-free navigation, avoiding both inter-robot and human-robot collisions. The results demonstrate the policy's robustness in handling dynamic, multi-agent environments with unpredictable disturbances.
	
	The navigation algorithm demonstrates robust performance in both single-agent and multi-agent scenarios. In the single-agent scenario, the TB3 successfully navigates through narrow passages and avoids obstacles of unknown positions and shapes, efficiently reaching the goal point. In a multi-agent scenario, the algorithm enables TB3s to avoid collisions with other robots and respond to dynamic disturbances, such as intentional human interference. The TB3s dynamically adjust their trajectories to accommodate sudden obstacles while maintaining collision-free navigation. These results highlight the algorithm's adaptability and reliability in handling static and dynamic challenges, ensuring safe and efficient navigation in real-world applications.
	
	\section{CONCLUSION}
	\label{conclusion}
	This paper presents LSTP-Nav, a lightweight end-to-end deep reinforcement learning framework that enables safe, real-time, map-free navigation for multi-robot systems operating in dynamic environments, using only raw LiDAR data and limited computational resources. Built upon the LSTP-Net architecture, the policy effectively extracts critical spatiotemporal features, addresses partial observability, and mitigates suboptimal behaviors such as freezing in narrow passages, while maintaining low computational overhead. A novel heading stability (HS) reward further improves collision avoidance by reducing unnecessary detours and enhancing stability in dense scenarios. To facilitate realistic training, we develop PhysReplay-Simlab, a physics-enhanced simulator that narrows the sim-to-real gap through realistic dynamics and targeted experience replay. Simulation and real-world experiments confirm the practicality and robustness of the proposed framework, demonstrating effective zero-shot sim-to-real transfer and real-time deployment on CPU-only platforms at over 40 Hz. Future work will explore extensions to dynamic traffic scenarios and UAV systems, refine the reward function to address complex edge cases, enhance adversarial robustness, and incorporate memory modules for improved long-term decision-making.
	
	\bibliographystyle{IEEEtran}
	\bibliography{reference.bib}
	
	\begin{IEEEbiography}
		[{\includegraphics[width=1in,height=1.25in,clip,keepaspectratio]{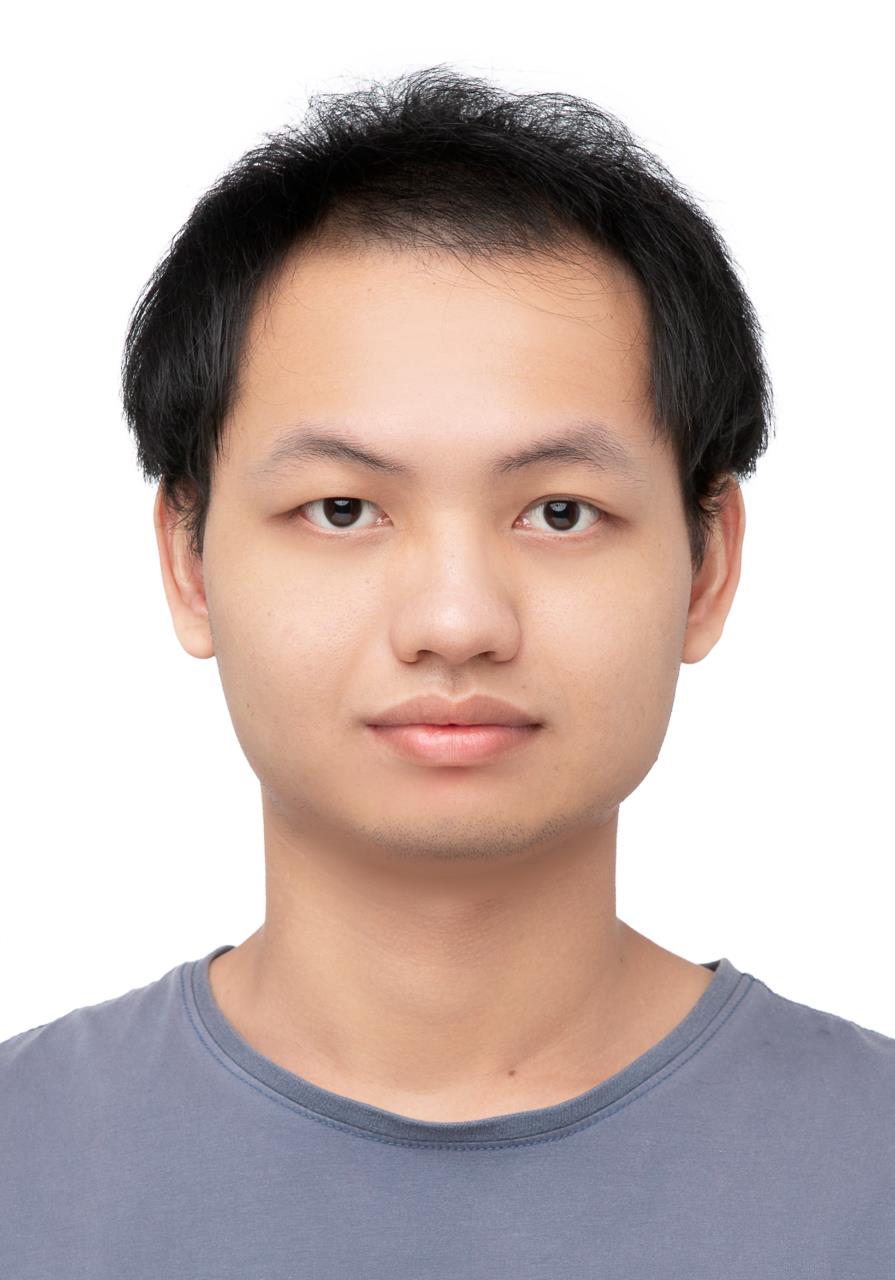}}] 
		{Xingrong Diao} received the B.E. degree in robotics engineering from the Mechanical and Energy Engineering, Southern University of Science and Technology, Shenzhen, China, in 2023. He is currently pursuing a Master's degree with the Department of Electronic and Electrical Engineering, Southern University of Science and Technology, Shenzhen, China. His research interests include robot manipulation and motion planning.
	\end{IEEEbiography}
		
	\begin{IEEEbiography}
		[{\includegraphics[width=1in,height=1.25in,clip,keepaspectratio]{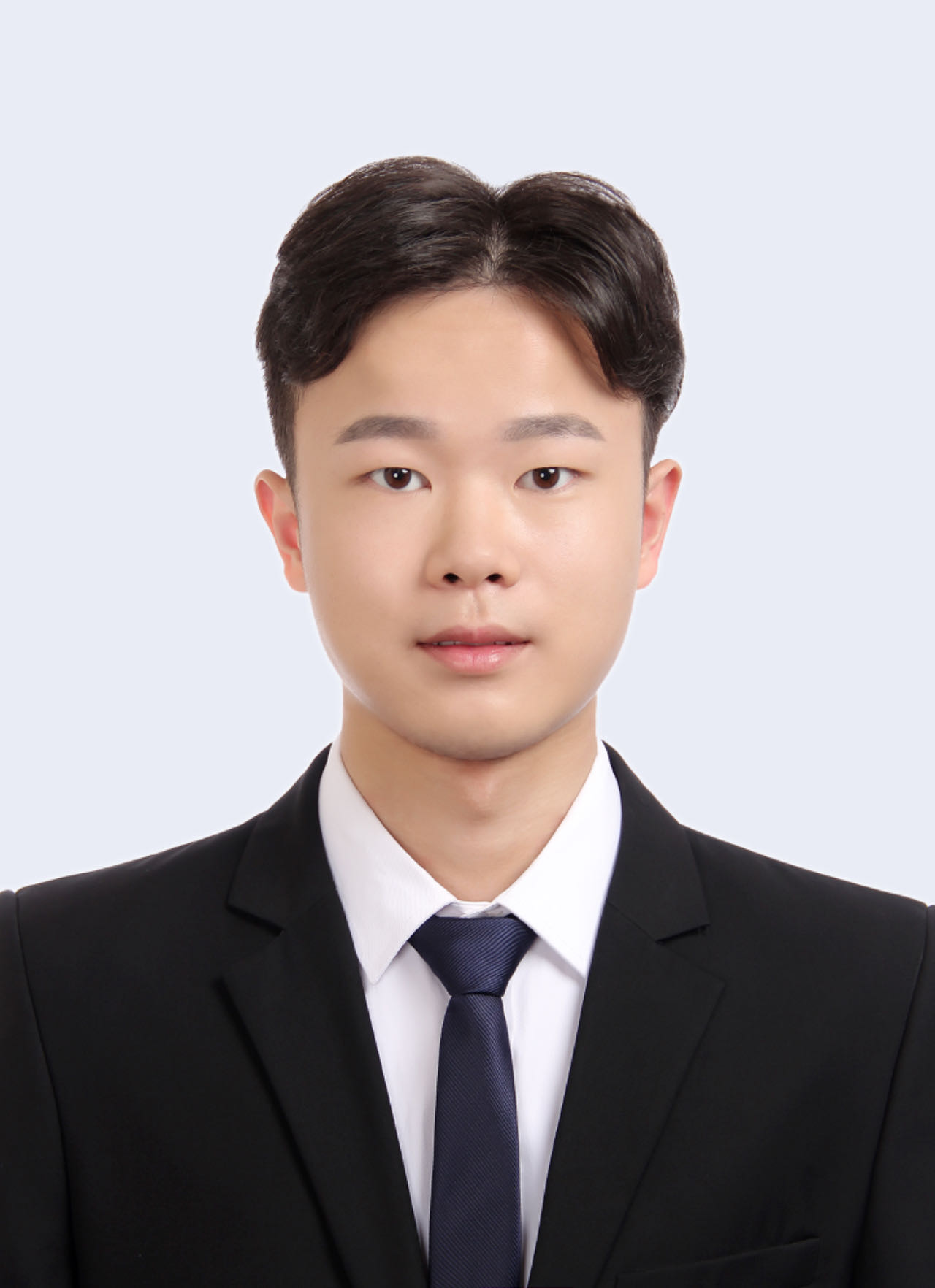}}] 
		{Zhirui Sun} received the B.E. degree in information engineering from the Department of Electronic and Electrical Engineering, Southern University of Science and Technology, Shenzhen, China, in 2019. He is currently pursuing the Ph.D. degree with the Department of Electronic and Electrical Engineering, Southern University of Science and Technology, Shenzhen, China. His research interests include robot perception and motion planning.
	\end{IEEEbiography}
	
	\begin{IEEEbiography}
		[{\includegraphics[width=1in,height=1.25in,clip,keepaspectratio]{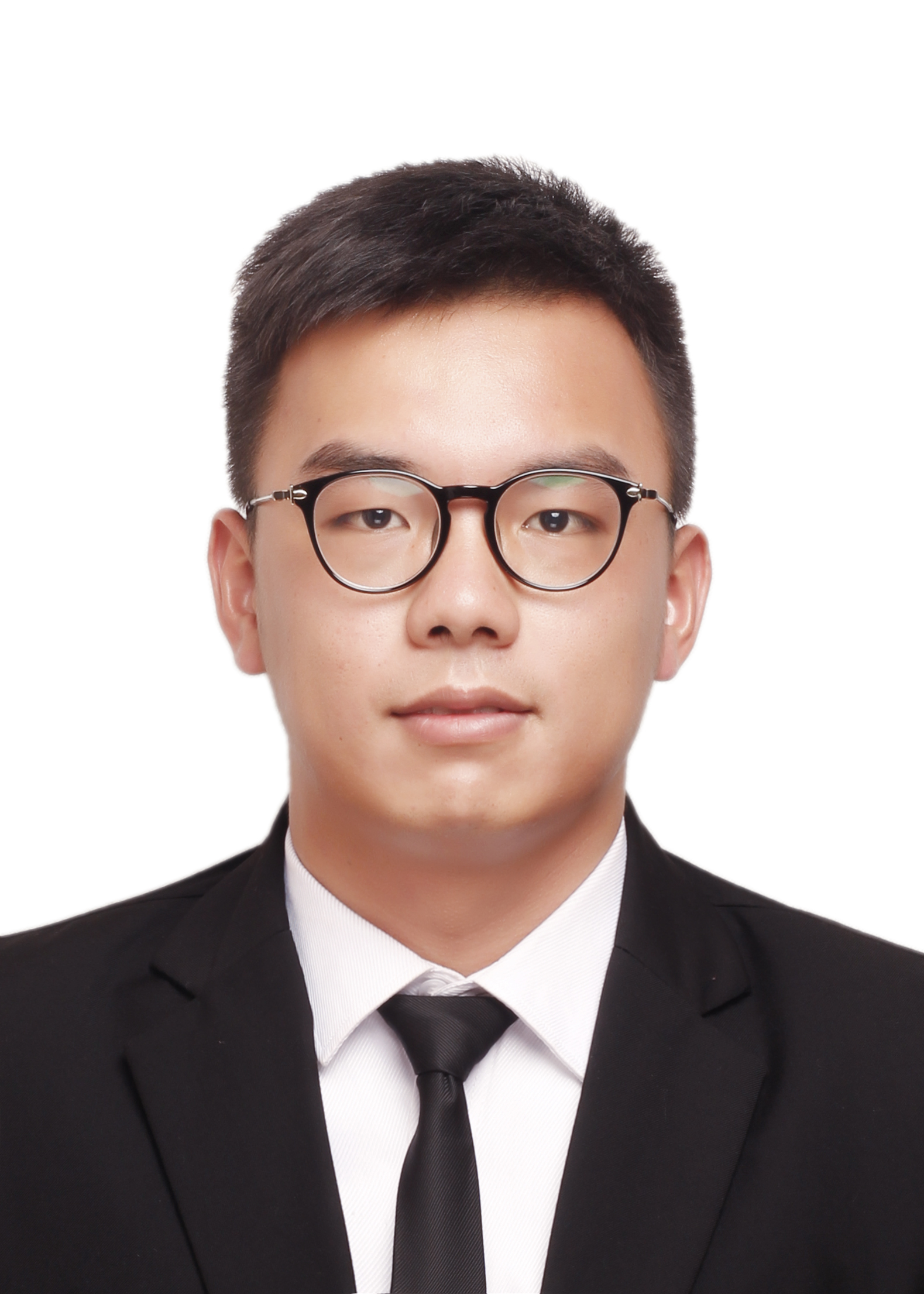}}]
		{Jianwei Peng} received the B.E. degree in automation from Huaqiao University, Xiamen, China, in 2020, and the M.E. degree in electronic and information engineering from the University of Chinese Academy of Sciences, Beijing, China, in 2023. He is currently pursuing the Ph.D. degree with the Department of Electronic and Electrical Engineering, Southern University of Science and Technology, Shenzhen, China. His research interests include motion planning, human-robot interaction, and social navigation.
	\end{IEEEbiography}
	
	\begin{IEEEbiography}
		[{\includegraphics[width=1in,height=1.25in,clip,keepaspectratio]{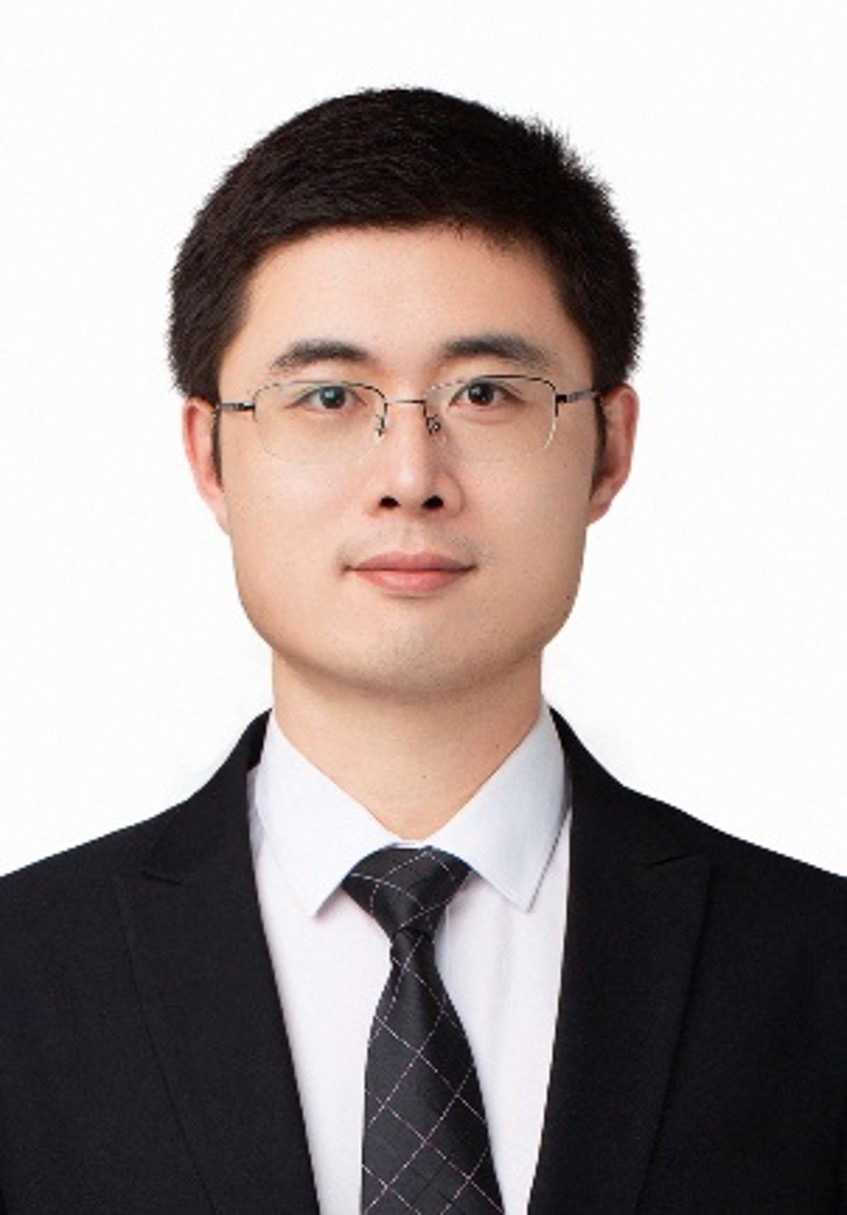}}] 
		{Jiankun Wang} (Senior Member, IEEE) received the B.E. degree in automation from Shandong University, Jinan, China, in 2015, and the Ph.D. degree from the Department of Electronic Engineering, The Chinese University of Hong Kong, Hong Kong, in 2019.
		
		He is currently an Assistant Professor at the Department of Electronic and Electrical Engineering, Southern University of Science and Technology, Shenzhen, China. His current research interests include motion planning and control, human-robot interaction, and machine learning in robotics.
		
		Currently, he serves as the associate editor of IEEE Transactions on Automation Science and Engineering, IEEE Transactions on Intelligent Vehicles, IEEE Robotics and Automation Letters, International Journal of Robotics and Automation, and Biomimetic Intelligence and Robotics.
	\end{IEEEbiography}
	
\end{document}